\documentclass[lettersize,journal]{IEEEtran}
\usepackage{amsmath,amsfonts,amssymb}
\usepackage{algorithm}
\usepackage{algpseudocodex}
\usepackage{array}
\usepackage{booktabs}
\usepackage{textcomp}
\usepackage{stfloats}
\usepackage{threeparttable}
\usepackage{multirow}
\usepackage{makecell}
\usepackage{tabularx}
\usepackage[table]{xcolor}
\usepackage{cite}
\usepackage{url}
\usepackage{verbatim}
\usepackage{graphicx}
\usepackage{xcolor}
\usepackage[colorlinks=true]{hyperref}
\hypersetup{
    bookmarks=false,
    pdfborder={0 0 0},
    linkcolor=blue,
    citecolor=blue,
    urlcolor=magenta!75!black
}
\newcolumntype{Y}{>{\centering\arraybackslash}X}

\begin{document}

\title{Elevator-LIO: Robust LiDAR-Inertial Odometry for Multi-Floor Navigation under Elevator-Induced Non-Inertial Motion}

\author{
Yifan~Zhang,
Yudong~Huang,
Yuchong~Zhang,
Changze~Li,
Haoran~Liu,
Ming~Yang,
and~Tong~Qin*
\\[0.5em]
Shanghai Jiao Tong University, Shanghai, China
}

\maketitle

\begin{abstract}
This paper presents Elevator-LIO, a LiDAR-inertial odometry framework designed to achieve continuous robot localization during elevator travel, thereby supporting cross-floor robotic tasks. To address the state-estimation problem in non-inertial frames, Elevator-LIO establishes a decoupled state-estimation model that separately models the robot motion relative to the elevator and the elevator motion itself, and embeds it into a mode-dependent iterated error-state Kalman filter framework. This framework degenerates to conventional LIO estimation in ordinary indoor environments, while enabling the propagation and constrained update of elevator-related states in elevator non-inertial environments, thereby achieving continuous and stable localization. An elevator mode manager detects elevator entry and exit events using LiDAR ranging statistics and estimated states, and introduces event-triggered zero-velocity and zero-acceleration updates when the elevator stops to suppress accumulated vertical drift. In addition, this paper adopts an adaptive voxel downsampling strategy to maintain a stable number of effective points under significant environmental scale changes. We conduct extensive experiments on 20 real-world sequences containing 79 elevator rides, including practical challenges such as large-scale spaces, long vertical travel, dynamic pedestrian interference, and mirror reflections. The results show that Elevator-LIO maintains continuous localization accuracy in all sequences, with terminal height error below 1 cm in 17 sequences. In contrast, existing representative localization systems perform poorly on these elevator sequences. Tests on the Hilti 2022/2023 datasets further show that the proposed method remains competitive in standard indoor scenarios. The project page is available at \url{https://xiaofan4122.github.io/Elevator_LIO_Page/}.
\end{abstract}

\begin{IEEEkeywords}
Multi-floor localization, LiDAR-inertial odometry, elevator travel, non-inertial environments, zero-velocity update
\end{IEEEkeywords}

\section{Introduction}
\IEEEPARstart{M}{ulti-floor} mobility is a fundamental requirement for autonomous mobile robots (AMRs) deployed in modern buildings, where elevators serve as the primary means of vertical transportation between floors. Cross-floor navigation via elevators is indispensable for practical indoor autonomy. However, it remains a major challenge for robot localization and frequently acts as a critical failure point in real-world deployments.

Over the past decade, vision-based SLAM systems \cite{vinsmono,orbslam3,droidslam} have achieved impressive accuracy but often struggle with illumination variations, motion blur, and high computational demands for dense mapping \cite{svo,dso}. LiDARs have been playing an increasingly important role in a wide range of autonomous robotic platforms, including autonomous ground vehicles (AGVs) navigating complex terrains \cite{legoloam} and unmanned aerial vehicles (UAVs) performing agile maneuvers in cluttered environments \cite{planning}.

Particularly, in GPS-denied indoor environments, LiDAR-inertial odometry (LIO) has become a mainstream solution for robot state estimation because it combines the geometric accuracy of LiDAR with the high-rate motion sensing of the IMU~\cite{xu2021fast_lio, fast_lio2, pointlio}. However, their success relies on a basic assumption: the sensor platform operates in an inertial reference frame, so that IMU measurements can be consistently interpreted as robot ego-motion and fused with LiDAR constraints.

\begin{figure*}[!t]
\centering
\includegraphics[width=\linewidth]{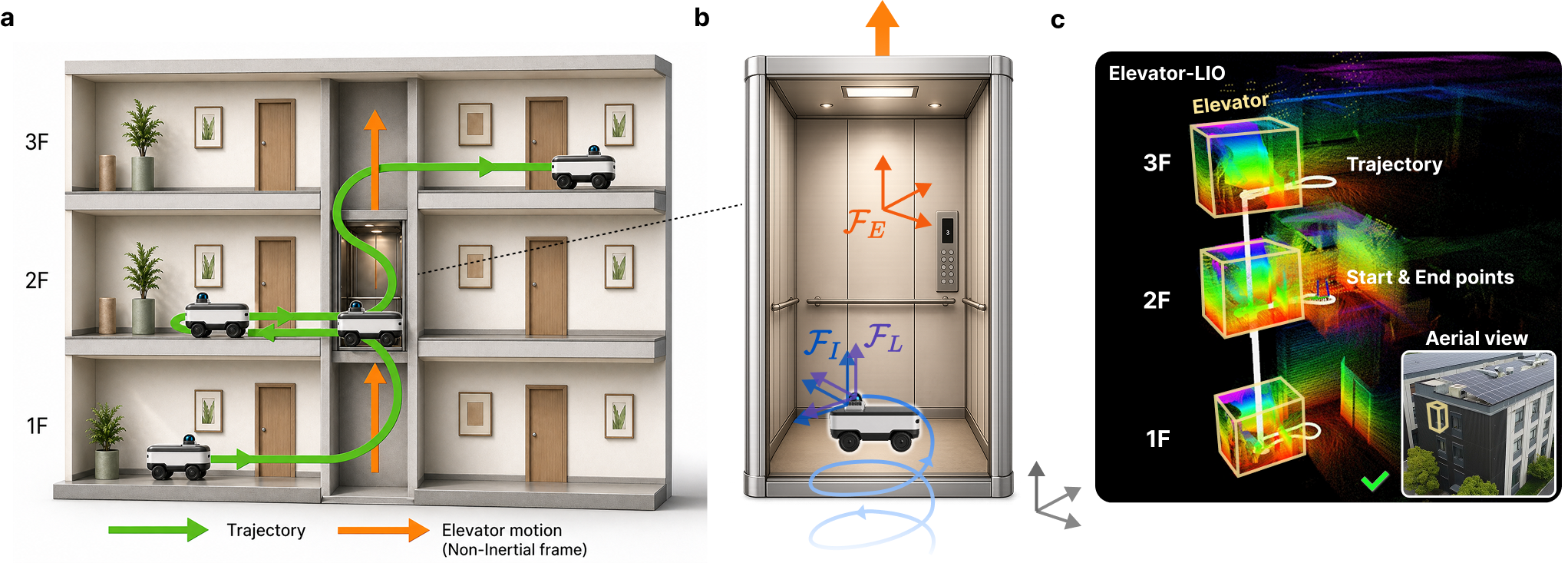}
\caption{
Conceptual illustration of Elevator-LIO. 
\textbf{a:} cross-floor robotic navigation requires continuous localization while the robot is transported by an elevator. 
\textbf{b:} inside the elevator cabin, the robot operates in a moving non-inertial frame, where the IMU captures motion affected by elevator transport, whereas the LiDAR observes cabin-relative geometry. 
Elevator-LIO addresses this sensing inconsistency by decoupling the robot-relative state from the elevator transport state. 
\textbf{c:} a representative real-world result showing continuous trajectory estimation and globally consistent multi-floor mapping across elevator traversal.
}
\label{fig:fig1_concept}
\end{figure*}

This assumption breaks down when the robot is inside a moving elevator. Once the robot enters the cabin, the platform is placed in a non-inertial environment, and the measured acceleration becomes contaminated by elevator motion. Meanwhile, LiDAR observations inside the cabin mainly provide local geometric constraints relative to the elevator interior rather than the world frame. Consequently, a conventional LIO pipeline may incorrectly attribute elevator-induced motion to the robot itself, leading to severe drift, inconsistency, or even divergence. This mismatch makes reliable state estimation during elevator motion a key bottleneck for multi-floor localization.

%~\cite{10886579}
%~\cite{10611182}

The difficulty arises from two aspects. First, from the estimation perspective, IMU measurements inside the cabin contain a superposition of elevator translational motion and robot motion relative to the elevator. Without explicitly separating these components, state propagation accumulates large integration errors. Moreover, since LiDAR can only observe relative motion inside the cabin, it cannot update the corresponding absolute-motion states. These states therefore become unobservable during elevator travel, and their uncertainty keeps increasing over time, making reliable estimation progressively more difficult. Second, from the front-end perspective, the robot experiences abrupt scene scale changes when moving between open floors and confined elevator cabins. Fixed-resolution downsampling then becomes unstable: overly dense observations impair real-time efficiency, whereas overly sparse observations can lead to rapid degradation or even failure in narrow elevator environments.

Related multi-floor or elevator-aware systems typically focus on mapping, navigation, or floor transition management \cite{zhao2020autonomous, jung2024munes, troniak2013charlie}, rather than maintaining continuous localization while the robot is physically inside a moving elevator. Therefore, a dedicated treatment of the non-inertial estimation problem is still missing.

To address these challenges, we propose \emph{Elevator-LIO}, a LiDAR-inertial localization framework for continuous operation across floor--elevator transitions. The key idea is to explicitly separate elevator motion from robot motion relative to the elevator and to estimate both within a unified state-estimation framework. By augmenting the elevator motion along the vertical direction into the system state and combining it with LiDAR relative constraints, the proposed method avoids directly treating elevator-induced acceleration as robot ego-motion, which is the main source of failure in conventional LIO during elevator traversal.

Because the elevator-related state remains weakly observable, we introduce a zero-state update (velocity and acceleration) mechanism at elevator stops to suppress accumulated vertical drift. To make this correction reliable in practice, we design a lightweight elevator state detector that identifies the onset and termination of elevator motion solely from onboard sensing, without relying on external infrastructure, thereby providing robust triggers for state transition and ZUPT correction.

In parallel, we develop a feedback-driven adaptive downsampling front-end to cope with the drastic change in observation scale between open floors and confined cabins. Instead of using a fixed voxel resolution, the front-end adjusts the downsampling scale online according to point-cloud feedback, so that the number of effective input points remains within a safe operating range. This design improves both computational stability and geometric conditioning across large scene transitions.

The proposed system operates using only a LiDAR with an integrated IMU. Experiments across public datasets, real-world data and a dedicated simulator show that Elevator-LIO maintains continuous, divergence-free localization during all tested elevator transitions, yields accurate floor-level vertical estimation, and preserves competitive performance in standard inertial environments.

The main contributions of this paper are summarized as follows:
\begin{enumerate}
    \item We propose a LiDAR--inertial localization framework for robot operation in elevators, and derive the decoupled motion equations in the elevator-induced non-inertial frame for stable estimation of both the robot pose relative to the elevator and its absolute pose.
    \item We develop a drift-suppression mechanism tailored to real elevator travel by combining zero state update correction with a lightweight elevator state detector, improving robustness against uncertainty accumulation during elevator motion.
    \item We introduce a feedback-driven adaptive downsampling strategy that maintains a consistent point cloud density, thereby ensuring robust front-end geometric constraints across abrupt scene-scale transitions between open floors and confined elevator cabins.
    \item We validate the proposed system through extensive experiments on public datasets, real-world elevator sequences and a dedicated simulator, and provide benchmark resources to support future research on non-inertial elevator localization.
\end{enumerate}

\section{Related Work}

Existing studies related to our problem mainly fall into two categories: LiDAR--inertial odometry (LIO) in inertial environments, and state estimation under moving-base or weakly observable conditions.

\begin{figure*}[!t]
\centering
\includegraphics[width=\linewidth]{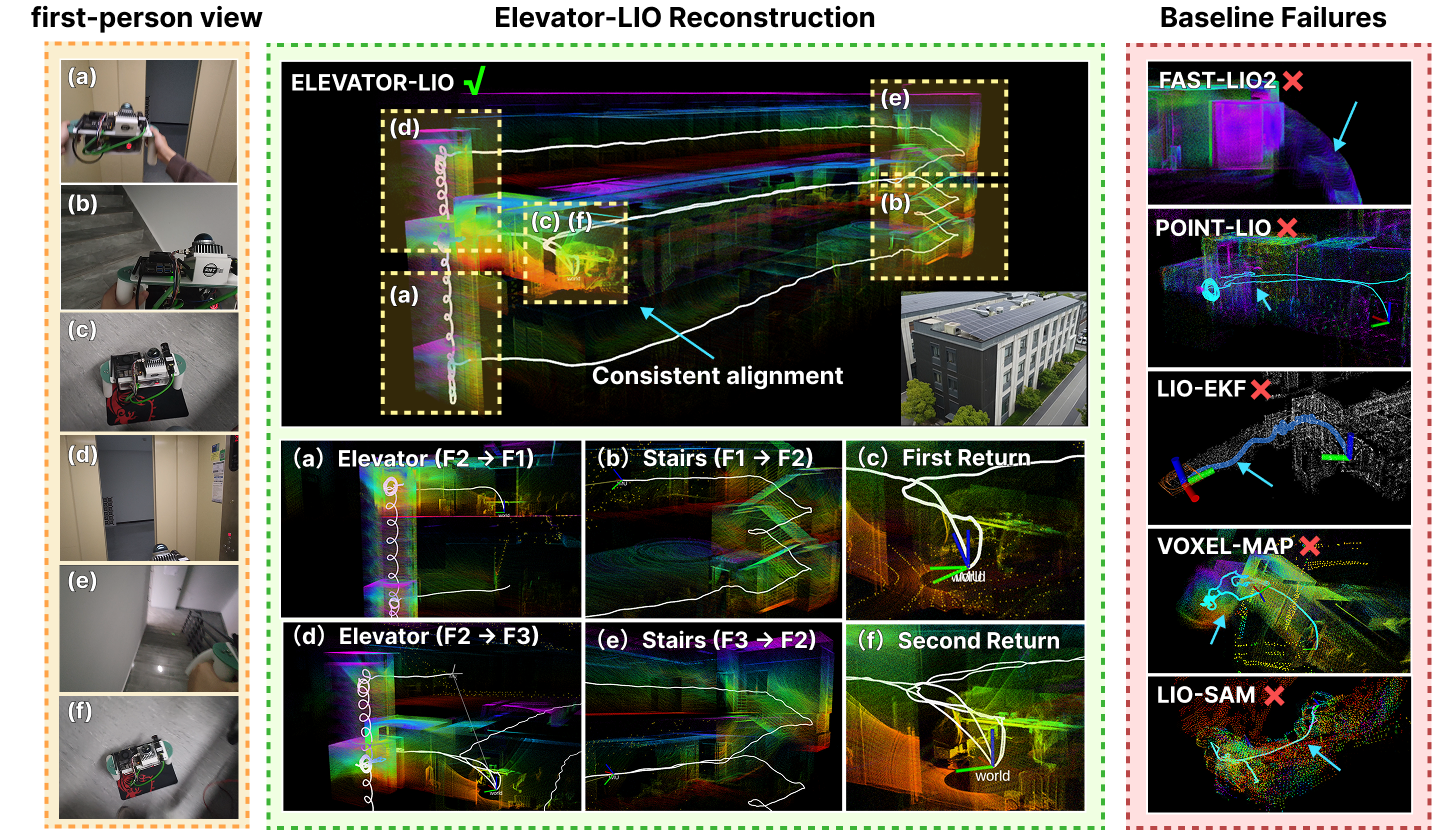}
\caption{Qualitative comparison of the handheld data collection sequence. The left column illustrates the first-person view alongside our sensor suite, capturing the complete trajectory with two consecutive revisits. The middle column shows that Elevator-LIO maintains continuous and geometrically consistent cross-floor localization. The local zoom-in views in (a)–(f) sequentially highlight our motion process. In (a) and (d), the robot undergoes aggressive motion inside the elevator, yet Elevator-LIO can still accurately separate and estimate the true motion components. Further experiments demonstrate that the baseline methods fail to maintain reliable localization even when the robot remains stationary inside the elevator.}
\label{fig:First_Figure}
\end{figure*}

\subsection{LiDAR--Inertial Odometry in Inertial Environments}

Modern LIO systems can be broadly grouped into smoothing-based methods, such as LIOM~\cite{ye2019tightly}, LIO-SAM~\cite{lio_sam}, and LILI-OM~\cite{liliom}, and filtering or direct-registration methods, such as FAST-LIO~\cite{xu2021fast_lio}, FAST-LIO2~\cite{fast_lio2}, Point-LIO~\cite{pointlio}, DLIO~\cite{chen2023dlio}, LIO-EKF~\cite{wu2024icra}, and Faster-LIO~\cite{fasterlio}. Later works further improve efficiency, map representation, and robustness through voxel maps, adaptive filtering, intensity cues, or additional gravity/velocity constraints~\cite{yuan2023voxelmapmergeablevoxelmapping,xie2025akf,10041769,pfreundschuh2024coin,cynoh-2025-icra}.

Despite these advances, standard LIO methods generally assume that the sensor platform evolves in an inertial frame, so that IMU measurements can be interpreted as robot ego-motion and fused consistently with LiDAR geometric constraints. This assumption breaks down inside a moving elevator: elevator-induced acceleration is superimposed on the IMU measurements, while LiDAR observations mainly constrain the robot relative to the cabin. As a result, the motion prior and geometric observations become physically inconsistent, making conventional LIO prone to vertical drift, map inconsistency, or divergence.

\subsection{State Estimation under Moving-Base or Non-Inertial Conditions}

State estimation under moving-base motion or weak observability often relies on additional constraints or external aiding. Ground-Fusion~\cite{yin2024ground_fusion} and VIWO~\cite{lee2020viwo} use wheel odometry or vehicle kinematic constraints, while other systems rely on external infrastructure to provide absolute observations. These assumptions are not well suited to elevator scenarios, where the robot may move within the cabin while the elevator undergoes significant vertical transport motion, and deployment should not depend on building-side infrastructure.

Another line of work exploits physically meaningful state-transition events. Zero-velocity updates have been widely used in inertial navigation~\cite{skog2010zupt}, and related contact or kinematic constraints have been incorporated into legged and multi-sensor odometry systems such as VILENS~\cite{wisth2023vilens}, Leg-KILO~\cite{legkilo2024}, and LIKO~\cite{zhao2024likolidarinertialkinematic}. These methods show that intermittent physical constraints can suppress drift under weak observability. However, they do not explicitly model the coupling between elevator transport motion and robot motion relative to the cabin.

Elevator motion also causes abrupt observation-scale changes between open floors and confined cabins. Prior work on degraded or adaptive LIO, such as LVIO-Fusion~\cite{zhang2024lvio_fusion} and AD-LIO~\cite{wang2026ad_lio}, improves robustness through multimodal observations or adaptive voxel resolution. In contrast, our work focuses on a dedicated LiDAR--inertial formulation for elevator-induced non-inertial motion, together with event-triggered constraints and adaptive front-end processing for continuous multi-floor localization.

\begin{figure*}[!t]
\centering
\includegraphics[width=1.00\textwidth]{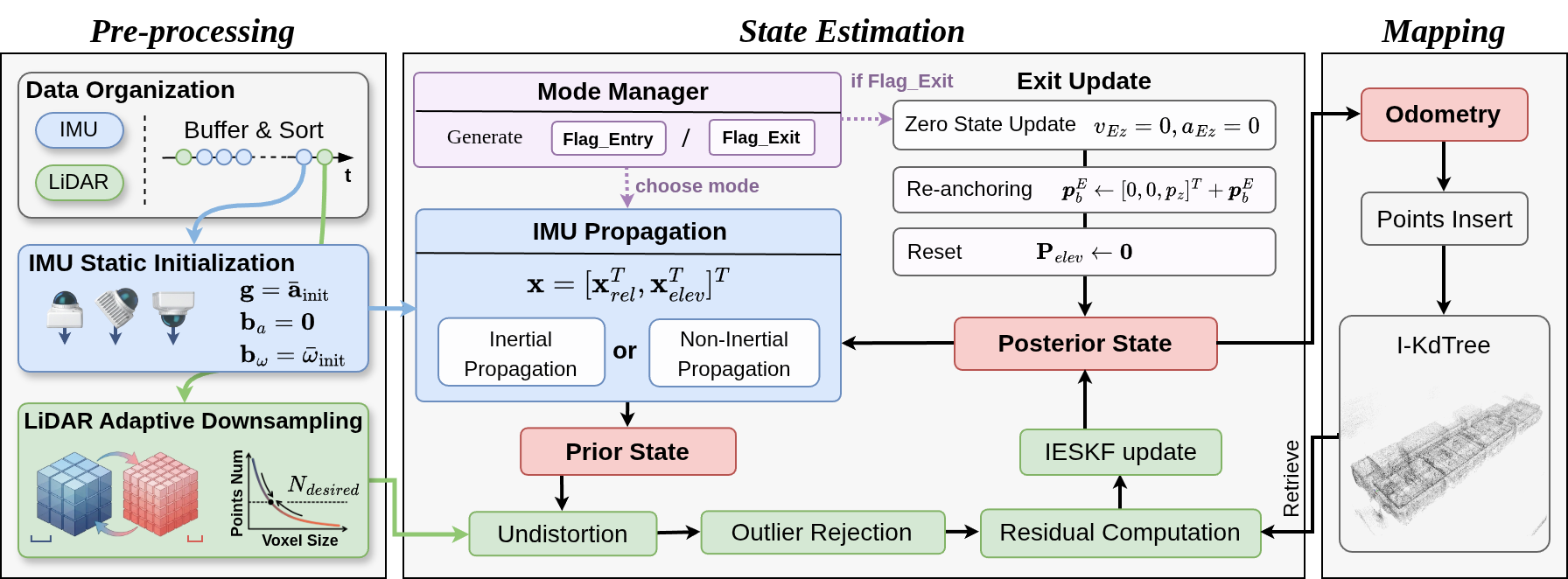}
\caption{Overview of the proposed Elevator-LIO framework. The system consists of pre-processing, elevator mode management, state estimation, and mapping.}
\label{fig:system_overview}
\end{figure*}

\section{System Overview}

As illustrated in Fig.~\ref{fig:system_overview}, Elevator-LIO consists of three main components: a pre-processing module, a state estimation module with elevator mode management, and a mapping module.

In the pre-processing stage, raw IMU and LiDAR measurements are first buffered and sorted by timestamp. A high-frequency timer then continuously polls the buffer and pops the earliest available sensor measurement for processing. The system then performs static IMU initialization using the initial stationary segment, including gravity alignment and bias estimation, so as to accommodate different sensor mounting configurations. Meanwhile, the raw point cloud is processed by a feedback-based adaptive downsampling module to maintain stable front-end inputs across environments with significantly different spatial scales, such as open floors and confined elevator cabins.

Unlike conventional LIO systems, Elevator-LIO re-derives the state-estimation formulation for elevator scenarios and elegantly integrates the resulting decoupled model with the original LIO estimation framework. During operation in ordinary inertial environments, the system behaves as a standard LIO pipeline. Once an elevator-induced non-inertial reference frame is detected, the estimator switches to the proposed non-inertial mode and explicitly distinguishes the motion of the moving reference frame from the robot's own motion. After leaving the non-inertial mode, an additional correction step, termed the \emph{Exit Update}, is applied to suppress accumulated errors and recover global-frame consistency.

Finally, the posterior state is published as the odometry output, while the retained points of the current scan are incrementally inserted into the global map maintained by an ikd-Tree.

\section{Notation and Preliminaries}

% This section first summarizes the notation used throughout the paper and then reviews standard inertial-frame IMU kinematics, which serves as the baseline for explaining why conventional LIO fails in elevator-induced non-inertial scenarios.

\subsection{Notation}
The key symbols used in Elevator-LIO are summarized in Table~\ref{tab:key_notations}. We use lowercase italic letters for scalars (e.g., $\Delta t$), lowercase bold letters for vectors (e.g., $\mathbf{p}$ and $\mathbf{v}$), and uppercase bold letters for matrices (e.g., $\mathbf{R}$ and $\mathbf{P}$).

Coordinate frames are denoted by calligraphic letters with subscripts.
For any frame $\mathcal{F}_A$ and reference frame $\mathcal{F}_B$,
$\mathbf{p}_A^B$ denotes the position of the origin of $\mathcal{F}_A$
with respect to $\mathcal{F}_B$, expressed in $\mathcal{F}_B$.
The corresponding orientation is represented by the Hamilton quaternion
$\mathbf{q}_A^B$ or the equivalent rotation matrix $\mathbf{R}_A^B$.

We also use $\lfloor \cdot \rfloor_\wedge$ for the skew-symmetric operator and $\mathrm{Exp}(\cdot)$ for the exponential map from $\mathfrak{so}(3)$ to $\mathrm{SO}(3)$.

\begin{table}[t]
\caption{Key notations used in Elevator-LIO.}
\label{tab:key_notations}
\centering
\renewcommand{\arraystretch}{1.20}
\setlength{\tabcolsep}{2pt}
\footnotesize
\begin{tabular*}{\columnwidth}{@{\extracolsep{\fill}}ll@{}}
\toprule
\textbf{Symbol} & \textbf{Definition} \\
\midrule
$\mathcal{F}_W,\mathcal{F}_E,\mathcal{F}_I$ & World, elevator-cabin, and IMU body frames \\
$\mathbf{p}_I^E$ & IMU position in $\mathcal{F}_E$ relative to elevator \\
$\mathbf{p}_E^W$ & Elevator position in $\mathcal{F}_W$ \\
$\mathbf{q}_A^B,\mathbf{R}_A^B$ & Quaternion and rotation from frame $A$ to frame $B$ \\
$\mathbf{v}_I^E$ & IMU velocity in $\mathcal{F}_E$ relative to elevator \\
$p_{Ez}^W, v_{Ez}^W, a_{Ez}^W$ & Elevator z displacement, velocity, and acceleration in $\mathcal{F}_W$ \\
$\mathbf{a}_m,\boldsymbol{\omega}_m$ & IMU specific-force and angular-rate measurements \\
${}^{I}\boldsymbol{\omega}_{I}^{E}$ & Angular velocity of $\mathcal{F}_I$ relative to $\mathcal{F}_E$, expressed in $\mathcal{F}_I$ \\
$\mathbf{b}_a,\mathbf{b}_{\omega}$ & Accelerometer and gyroscope biases \\
$\mathbf{g}$ & Gravity vector in $\mathcal{F}_W$ \\
$\mathbf{e}_3$ & Unit z-axis vector, i.e., $[0,0,1]^T$ \\
$\lfloor\cdot\rfloor_\wedge$ & Skew operator: $\mathbb{R}^3 \rightarrow \mathfrak{so}(3)$ \\
\bottomrule
\end{tabular*}
\end{table}

\subsection{Standard IMU Kinematics}
% With the above notation, we briefly revisit standard inertial-frame IMU kinematics, as it forms the basis of conventional LIO methods and highlights why such formulations fail in elevator scenarios.

In classical LIO, the world frame $\mathcal{F}_W$ is assumed to be inertial. Under this assumption, the system state can be written as
\begin{equation}
\mathbf{x}_{std} =
\begin{bmatrix}
(\mathbf{p}_I^W)^T &
(\mathbf{q}_I^W)^T &
(\mathbf{v}_I^W)^T &
\mathbf{b}_a^T &
\mathbf{b}_\omega^T &
\mathbf{g}^T
\end{bmatrix}^T
\end{equation}
% where $\mathbf{b}_a$ and $\mathbf{b}_\omega$ denote the accelerometer and gyroscope biases, respectively, and $\mathbf{g}$ is the gravity vector expressed in the world frame.

% Let $\boldsymbol{\omega}_m$ and $\mathbf{a}_m$ denote the measured angular velocity and linear acceleration from the IMU. 
The corresponding continuous-time dynamics under the inertial-frame assumption are given by
\begin{equation}
\left\{
\begin{aligned}
\dot{\mathbf{p}}_I^W &= \mathbf{v}_I^W, \\
\dot{\mathbf{q}}_I^W &= \mathbf{q}_I^W \otimes
\begin{bmatrix}
0 \\
\frac{1}{2}\left(\boldsymbol{\omega}_m - \mathbf{b}_\omega - \mathbf{n}_\omega\right)
\end{bmatrix}, \\
\dot{\mathbf{v}}_I^W &= \mathbf{R}_I^W \left(\mathbf{a}_m - \mathbf{b}_a - \mathbf{n}_a\right) + \mathbf{g}, \\
\dot{\mathbf{b}}_a &= \mathbf{n}_{ba}, \qquad
\dot{\mathbf{b}}_\omega = \mathbf{n}_{b\omega},
\end{aligned}
\right.
\end{equation}
where $\mathbf{n}_\omega$ and $\mathbf{n}_a$ are Gaussian white noises associated with the gyroscope and accelerometer measurements, respectively, and $\mathbf{n}_{ba}$ and $\mathbf{n}_{b\omega}$ represent the corresponding bias random walks.

The above model is valid only when the sensor platform evolves in an inertial frame. However, when the robot is carried by a moving elevator, the measured acceleration $\mathbf{a}_m$ contains both the robot's relative motion and the elevator transport motion. Elevator-LIO addresses this limitation through the non-inertial formulation presented in the next section.

\section{Elevator-LIO Framework}
This section presents the complete framework of Elevator-LIO shown in Fig.~\ref{fig:system_overview}. We begin with the core estimator, including the non-inertial state representation and its elevator-specific simplification (\ref{subsec:non_inertial_state_representation}), the corresponding propagation and covariance model for both non-inertial (\ref{subsec:non_inertial_state_propagation}) and inertial cases (\ref{subsec:inertial_state_propagation}), and the LiDAR update together with the elevator-exit correction (\ref{subsec:state_update_and_exit_constraint}). We then introduce several supporting modules that are equally important to the practical operation of the system, including the elevator mode manager (\ref{subsec:elevator_mode_manager}), static initialization (\ref{subsec:static_initialization}), geometry-aware adaptive downsampling (\ref{subsec:adaptive_downsampling}), and mapping (\ref{subsec:mapping}).

\subsection{Non-Inertial State Representation}
\label{subsec:non_inertial_state_representation}

Conventional tightly coupled LiDAR--IMU odometry typically parameterizes the robot state directly in an inertial world frame. In non-inertial scenarios, however, such a formulation becomes inconsistent with the actual motion composition. To account for this effect, we introduce an augmented state defined with respect to a generalized moving non-inertial frame $\mathcal{F}_N$, which decomposes the motion into the robot motion relative to $\mathcal{F}_N$ and the motion of $\mathcal{F}_N$ in the world frame. In the specific elevator scenario considered in this paper:
\begin{equation}
\mathbf{x}_{aug} =
\begin{bmatrix}
\mathbf{x}_{rel}^T & \mathbf{x}_{N}^T
\end{bmatrix}^T
\end{equation}
with
\begin{equation}
\mathbf{x}_{rel} =
\begin{bmatrix}
(\mathbf{p}_I^{N})^T &
(\mathbf{q}_I^{N})^T &
(\mathbf{v}_I^{N})^T &
\mathbf{b}_a^T &
\mathbf{b}_\omega^T &
\mathbf{g}^T
\end{bmatrix}^T
\end{equation}
and
\begin{equation}
\mathbf{x}_{N} =
\begin{bmatrix}
(\mathbf{p}_N^W)^T &
(\mathbf{q}_N^W)^T &
(\mathbf{v}_N^W)^T
\end{bmatrix}^T.
\end{equation}

Compared with conventional LIO, the robot state is no longer expressed directly in $\mathcal{F}_W$; instead, the transport motion of the non-inertial frame is modeled explicitly.

\subsubsection{Generalized Kinematic Propagation}

According to the relative motion theorem, the IMU kinematics in the world frame satisfy
\begin{equation}
\mathbf{R}_I^W = \mathbf{R}_N^W \mathbf{R}_I^N
\end{equation}
and
\begin{equation}
\left\{
\begin{aligned}
\mathbf{p}_I^W &= \mathbf{p}_N^W + \mathbf{R}_N^W \mathbf{p}_I^N, \\
\mathbf{v}_I^W &= \mathbf{v}_N^W + \mathbf{R}_N^W \mathbf{v}_I^N
+ \boldsymbol{\omega}_N^W \times \left(\mathbf{R}_N^W \mathbf{p}_I^N\right), \\
\mathbf{a}_I^W &= \mathbf{a}_N^W + \mathbf{R}_N^W \mathbf{a}_I^N
+ 2\boldsymbol{\omega}_N^W \times \left(\mathbf{R}_N^W \mathbf{v}_I^N\right) \\
&\quad + \dot{\boldsymbol{\omega}}_N^W \times \left(\mathbf{R}_N^W \mathbf{p}_I^N\right)
+ \boldsymbol{\omega}_N^W \times \left(\boldsymbol{\omega}_N^W \times \left(\mathbf{R}_N^W \mathbf{p}_I^N\right)\right),
\end{aligned}
\right.
\end{equation}
where $\boldsymbol{\omega}_N^W$ denotes the angular velocity of the non-inertial frame with respect to the world frame. The acceleration terms correspond to transport acceleration, relative acceleration, Coriolis acceleration, Euler acceleration, and centripetal acceleration, respectively.

The above formulation describes the complete motion composition in a general non-inertial setting. The actual estimator implemented in Elevator-LIO adopts a reduced parameterization tailored to elevator motion.

\subsubsection{Elevator-Specific Simplification}

At static initialization, the world frame $\mathcal{F}_W$ is defined with its $z$-axis aligned with gravity. In the elevator scenario considered in this paper, the cabin undergoes vertical translation during a single elevator traversal, while its rotational motion is negligible. Therefore, the elevator orientation relative to $\mathcal{F}_W$ can be regarded as constant, i.e.,
\begin{equation}
\boldsymbol{\omega}_E^W \approx \mathbf{0}.
\end{equation}
The elevator frame $\mathcal{F}_E$ is defined to remain parallel to the world frame $\mathcal{F}_W$. Therefore,
\begin{equation}
\mathbf{R}_E^W = \mathbf{I}.
\end{equation}
Under this frame choice, the transport motion of the elevator reduces to pure translation along the vertical axis, and the acceleration composition simplifies to
\begin{equation}
\mathbf{a}_I^W
=
\mathbf{a}_E^W + \mathbf{R}_E^W \mathbf{a}_I^E
=
\begin{bmatrix}
0 & 0 & a_{Ez}^W
\end{bmatrix}^T
+ \mathbf{a}_I^E.
\end{equation}

To obtain a compact yet sufficient parameterization, constant and redundant terms are removed from the state representation. The final state adopted in Elevator-LIO is
\begin{equation}
\mathbf{x} =
\begin{bmatrix}
\mathbf{x}_{rel}^T & \mathbf{x}_{elev}^T
\end{bmatrix}^T
\end{equation}
with
\begin{equation}
\mathbf{x}_{rel} =
\begin{bmatrix}
(\mathbf{p}_I^{E})^T &
(\mathbf{q}_I^{E})^T &
(\mathbf{v}_I^{E})^T &
\mathbf{b}_a^T &
\mathbf{b}_\omega^T
\end{bmatrix}^T
\end{equation}
and
\begin{equation}
\mathbf{x}_{elev} =
\begin{bmatrix}
p_{Ez}^W &
v_{Ez}^W &
a_{Ez}^W
\end{bmatrix}^T
\end{equation}

Here, $p_{Ez}^W$, $v_{Ez}^W$, and $a_{Ez}^W$ denote the elevator displacement, velocity, and acceleration along the world-frame vertical axis, respectively. Among these transport states, the vertical velocity and position accumulate errors during in-elevator motion and is further corrected by the exit update introduced later. Since the elevator rotation is assumed negligible, the attitude evolution is entirely represented by the robot orientation relative to the elevator frame.

\subsubsection{Design rationale}
We explicitly include $a_{Ez}^W$ in the state because recovering the robot's true motion requires explicitly estimating the motion of the reference frame. We choose to model the elevator acceleration $a_{Ez}^W$, rather than the robot-relative acceleration $a_{Iz}^E$, because the former usually varies more smoothly during elevator operation. A somewhat counter-intuitive aspect is that the unmodeled component is immediately interpreted by the IMU propagation, whereas the modeled component is treated as a latent state with random-walk dynamics and is mainly corrected through LiDAR updates and exit-stage constraints. Since LiDAR updates are relatively low-frequency compared with IMU propagation, it is preferable to assign the smoother component to the estimated state. Therefore, Elevator-LIO models $a_{Ez}^W$ as a random walk and estimates it online.

Introducing higher-order transport dynamics, such as jerk, would increase the state dimension and noise sensitivity, thereby degrading numerical robustness. The adopted parameterization therefore provides a trade-off between representational adequacy and estimator stability.

In addition, gravity is fixed after static initialization rather than estimated online. In the elevator frame, gravity becomes strongly coupled with IMU bias, elevator transport acceleration, and robot-relative acceleration. Estimating gravity online would therefore aggravate parameter coupling and reduce numerical stability during the weakly observable in-elevator phase. We thus keep gravity fixed and let the remaining vertical transport effect be captured by the explicitly modeled elevator-motion states.

\subsection{Non-Inertial State Propagation}
\label{subsec:non_inertial_state_propagation}

\subsubsection{IMU Measurement Model and Discrete-Time Kinematics}

Since the elevator rotational motion is neglected, the angular-velocity composition reduces to 

\begin{equation}
{}^{I}\boldsymbol{\omega}_{I}^W =  {}^{I}\boldsymbol{\omega}_{E}^W + {}^{I}\boldsymbol{\omega}_{I}^E \approx {}^{I}\boldsymbol{\omega}_{I}^E
\end{equation}

Here, the left superscript $I$ indicates that all angular velocities in this equation are expressed in the IMU frame $\mathcal{F}_I$. Let $\boldsymbol{\omega}_m$ denote the gyroscope measurement and $\mathbf{a}_m$ the accelerometer specific-force measurement. The IMU measurement model under the non-inertial elevator formulation is then written as
\begin{equation}
\left\{
\begin{aligned}
\boldsymbol{\omega}_m
&=
{}^{I}\boldsymbol{\omega}_I^E + \mathbf{b}_\omega + \mathbf{n}_\omega, \\
\mathbf{a}_m
&=
\left(\mathbf{R}_I^E\right)^T
\left(
\mathbf{a}_I^E + a_{Ez}^W \mathbf{e}_3 - \mathbf{g}
\right)
+ \mathbf{b}_a + \mathbf{n}_a,
\end{aligned}
\right.
\end{equation}
where $\mathbf{n}_\omega$ and $\mathbf{n}_a$ are the Gaussian white noises of the gyroscope and accelerometer, respectively. Since the elevator frame is initialized to be aligned with the world frame during static initialization, the gravity vector $\mathbf{g}$ and the vertical unit vector $\mathbf{e}_3=[0,0,1]^T$ have identical coordinate representations in $\mathcal{F}_E$ and $\mathcal{F}_W$.

For discrete-time filtering, the above model is discretized over the sampling interval $\Delta t$ between time steps $k$ and $k+1$. Define the effective relative acceleration in the elevator frame as
\begin{equation}
\mathbf{a}_{I,k}^E
=
\mathbf{R}_{I,k}^E \left(\mathbf{a}_{m,k} - \mathbf{b}_{a,k}\right)
+ \mathbf{g}
- a_{Ez,k}^W \mathbf{e}_3.
\end{equation}
The corresponding nominal-state propagation equations are
\begin{equation}
\label{eq:non_inert}
\left\{
\begin{aligned}
\mathbf{p}_{I,k+1}^E
&=
\mathbf{p}_{I,k}^E + \mathbf{v}_{I,k}^E \Delta t
+ \frac{1}{2}\mathbf{a}_{I,k}^E \Delta t^2, \\
\mathbf{q}_{I,k+1}^E
&=
\mathbf{q}_{I,k}^E \otimes
\mathrm{Exp}\!\left(
\left(\boldsymbol{\omega}_{m,k} - \mathbf{b}_{\omega,k}\right)\Delta t
\right), \\
\mathbf{v}_{I,k+1}^E
&=
\mathbf{v}_{I,k}^E + \mathbf{a}_{I,k}^E \Delta t, \\
\mathbf{b}_{a,k+1}
&=
\mathbf{b}_{a,k}, \\
\mathbf{b}_{\omega,k+1}
&=
\mathbf{b}_{\omega,k}, \\
p_{Ez,k+1}^W
&=
p_{Ez,k}^W + v_{Ez,k}^W \Delta t + \frac{1}{2} a_{Ez,k}^W \Delta t^2, \\
v_{Ez,k+1}^W
&=
v_{Ez,k}^W + a_{Ez,k}^W \Delta t, \\
a_{Ez,k+1}^W
&=
a_{Ez,k}^W.
\end{aligned}
\right.
\end{equation}

In the nominal propagation, the accelerometer bias, gyroscope bias, and elevator acceleration are treated as constant over one integration interval. Their uncertainties are modeled as random walks and injected through the process-noise covariance in the error-state propagation described next.

\subsubsection{Error-State Covariance Propagation}

The error state is defined as
\begin{equation}
\delta \mathbf{x} =
\begin{bmatrix}
\delta \mathbf{x}_r^T & \delta \mathbf{x}_e^T
\end{bmatrix}^T
\in \mathbb{R}^{18},
\end{equation}
where
\begin{equation}
\delta \mathbf{x}_r =
\begin{bmatrix}
(\delta \mathbf{p}_I^E)^T &
(\delta \boldsymbol{\theta}_I^E)^T &
(\delta \mathbf{v}_I^E)^T &
\delta \mathbf{b}_a^T &
\delta \mathbf{b}_\omega^T
\end{bmatrix}^T
\in \mathbb{R}^{15},
\end{equation}
and
\begin{equation}
\delta \mathbf{x}_e =
\begin{bmatrix}
\delta p_{Ez}^W &
\delta v_{Ez}^W &
\delta a_{Ez}^W
\end{bmatrix}^T
\in \mathbb{R}^{3}.
\end{equation}
For brevity, the bias-corrected measurements are denoted by
\begin{equation}
\hat{\boldsymbol{\omega}}_k = \boldsymbol{\omega}_{m,k} - \mathbf{b}_{\omega,k},
\qquad
\hat{\mathbf{a}}_k = \mathbf{a}_{m,k} - \mathbf{b}_{a,k}.
\end{equation}
For covariance propagation, the linearized error-state dynamics are written in discrete time as
\begin{equation}
\label{eq:delta_x_propagation}
\delta \mathbf{x}_{k+1}
=
\mathbf{F}_{x,k}\,\delta \mathbf{x}_k
+
\mathbf{F}_{w,k}\,\mathbf{w}_k,
\end{equation}
where $\mathbf{F}_{x,k}$ is the discrete-time error-state transition matrix, as shown in \eqref{eq:Fx},
$\mathbf{F}_{w,k}$ is the discrete-time noise injection matrix, and
$\mathbf{w}_k$ is the discretized process noise over one propagation interval.
We model
\begin{equation}
\label{eq:wk_distribution}
\mathbf{w}_k \sim \mathcal{N}(\mathbf{0}, \mathbf{Q}_{w,k}),
\end{equation}
with
\begin{equation}
\label{eq:Qw}
\mathbf{Q}_{w,k}
=
\mathrm{diag}
\!\left(
\sigma_a^2 \,\mathbf{I}_3,\;
\sigma_\omega^2 \,\mathbf{I}_3,\;
\sigma_{ba}^2 \,\mathbf{I}_3,\;
\sigma_{b\omega}^2 \,\mathbf{I}_3,\;
\sigma_{aE}^2 
\right)\Delta t
\end{equation}
Accordingly, the covariance is propagated as
\begin{equation}
\label{eq:P_propagation}
\mathbf{P}_{k+1}
=
\mathbf{F}_{x,k}\,\mathbf{P}_k\,\mathbf{F}_{x,k}^T
+
\mathbf{F}_{w,k}\,\mathbf{Q}_{w,k}\,\mathbf{F}_{w,k}^T.
\end{equation}

\begin{figure*}[!t]
\centering
\setlength{\arraycolsep}{4pt}
\begin{equation}
\label{eq:Fx}
\mathbf{F}_{x,k} =
\left[
\begin{array}{ccccc|ccc}
\mathbf{I}_3 &
-\frac{1}{2}\mathbf{R}_{I,k}^E
\lfloor \hat{\mathbf{a}}_k \rfloor_\wedge \Delta t^2 &
\mathbf{I}_3 \Delta t &
-\frac{1}{2}\mathbf{R}_{I,k}^E \Delta t^2 &
\mathbf{0} &
\mathbf{0}_{3\times 1} &
\mathbf{0}_{3\times 1} &
-\frac{1}{2}\mathbf{e}_3 \Delta t^2 \\

\mathbf{0} &
\mathrm{Exp}(-\hat{\boldsymbol{\omega}}_k \Delta t) &
\mathbf{0} &
\mathbf{0} &
-\mathbf{I}_3 \Delta t &
\mathbf{0}_{3\times 1} &
\mathbf{0}_{3\times 1} &
\mathbf{0}_{3\times 1} \\

\mathbf{0} &
-\mathbf{R}_{I,k}^E \lfloor \hat{\mathbf{a}}_k \rfloor_\wedge \Delta t &
\mathbf{I}_3 &
-\mathbf{R}_{I,k}^E \Delta t &
\mathbf{0} &
\mathbf{0}_{3\times 1} &
\mathbf{0}_{3\times 1} &
-\mathbf{e}_3 \Delta t \\

\mathbf{0} &
\mathbf{0} &
\mathbf{0} &
\mathbf{I}_3 &
\mathbf{0} &
\mathbf{0}_{3\times 1} &
\mathbf{0}_{3\times 1} &
\mathbf{0}_{3\times 1} \\

\mathbf{0} &
\mathbf{0} &
\mathbf{0} &
\mathbf{0} &
\mathbf{I}_3 &
\mathbf{0}_{3\times 1} &
\mathbf{0}_{3\times 1} &
\mathbf{0}_{3\times 1} \\
\hline
\mathbf{0}_{1\times 3} &
\mathbf{0}_{1\times 3} &
\mathbf{0}_{1\times 3} &
\mathbf{0}_{1\times 3} &
\mathbf{0}_{1\times 3} &
1 &
\Delta t &
\frac{1}{2}\Delta t^2 \\

\mathbf{0}_{1\times 3} &
\mathbf{0}_{1\times 3} &
\mathbf{0}_{1\times 3} &
\mathbf{0}_{1\times 3} &
\mathbf{0}_{1\times 3} &
0 &
1 &
\Delta t \\

\mathbf{0}_{1\times 3} &
\mathbf{0}_{1\times 3} &
\mathbf{0}_{1\times 3} &
\mathbf{0}_{1\times 3} &
\mathbf{0}_{1\times 3} &
0 &
0 &
1
\end{array}
\right]
\end{equation}
\end{figure*}
The corresponding discrete-time noise injection matrix
$\mathbf{F}_{w,k}\in\mathbb{R}^{18\times13}$ is given by
\begin{equation}
\label{eq:Fw}
\mathbf{F}_{w,k} =
\left[
\begin{array}{cccc|c}
\mathbf{0}_3 & \mathbf{0}_3 & \mathbf{0}_3 & \mathbf{0}_3 & \mathbf{0}_{3\times 1} \\
\mathbf{0}_3 & -\mathbf{I}_3 & \mathbf{0}_3 & \mathbf{0}_3 & \mathbf{0}_{3\times 1} \\
-\mathbf{R}_{I,k}^E & \mathbf{0}_3 & \mathbf{0}_3 & \mathbf{0}_3 & \mathbf{0}_{3\times 1} \\
\mathbf{0}_3 & \mathbf{0}_3 & \mathbf{I}_3 & \mathbf{0}_3 & \mathbf{0}_{3\times 1} \\
\mathbf{0}_3 & \mathbf{0}_3 & \mathbf{0}_3 & \mathbf{I}_3 & \mathbf{0}_{3\times 1} \\
\hline
\mathbf{0}_{1\times 3} & \mathbf{0}_{1\times 3} & \mathbf{0}_{1\times 3} & \mathbf{0}_{1\times 3} & 0 \\
\mathbf{0}_{1\times 3} & \mathbf{0}_{1\times 3} & \mathbf{0}_{1\times 3} & \mathbf{0}_{1\times 3} & 0 \\
\mathbf{0}_{1\times 3} & \mathbf{0}_{1\times 3} & \mathbf{0}_{1\times 3} & \mathbf{0}_{1\times 3} & 1
\end{array}
\right]
\end{equation}

Here, $\sigma_a^2$, $\sigma_\omega^2$, $\sigma_{ba}^2$, $\sigma_{b\omega}^2$, and
$\sigma_{aE}^2$ denote the continuous-time noise intensities of the accelerometer
measurement noise, gyroscope measurement noise, accelerometer-bias random walk,
gyroscope-bias random walk, and elevator-acceleration random walk, respectively.

\begin{algorithm}[t]
\caption{State Estimation Pipeline of Elevator-LIO}
\label{alg:elevator_lio}
\small
\begin{algorithmic}[1]
\Require LiDAR scan $\mathbf{z}_k$ or IMU frame $\mathcal{U}_k$ in buffer
\State Pop the oldest measurement from buffer
\If{measurement is IMU}
    \Statex \textcolor{gray}{// IMU propagation}
    \State Propagate state $\bar{\mathbf{x}}_k$ via \eqref{eq:non_inert}
    \State Propagate covariance $\mathbf{P}_k$ via \eqref{eq:P_propagation}
\ElsIf{measurement is LiDAR}
    \Statex \textcolor{gray}{// LiDAR update}
    \State Downsample, undistort, and reject outlier points
    \State $\kappa \gets 0$, $\hat{\mathbf{x}}_k^{(0)} \gets \bar{\mathbf{x}}_k$
    \While{$\kappa < \kappa_{\max}$}
        \State Compute residual via \eqref{eq:residual}
        \State Compute state increment $\delta \mathbf{x}_k^{(\kappa)}$ via \eqref{eq:update_x}
        \State $\hat{\mathbf{x}}_k^{(\kappa+1)} \gets \hat{\mathbf{x}}_k^{(\kappa)} \boxplus \delta \mathbf{x}_k^{(\kappa)}$
        \If{$\left\|\hat{\mathbf{x}}_k^{(\kappa+1)} \boxminus \hat{\mathbf{x}}_k^{(\kappa)}\right\| < \varepsilon$}
            \State \textbf{break}
        \EndIf
        \State $\kappa \gets \kappa + 1$
    \EndWhile
    \State $\hat{\mathbf{x}}_k \gets \hat{\mathbf{x}}_k^{(\kappa+1)}$
    \State Update covariance $\mathbf{P}_k$ via \eqref{eq:update_P}

    \Statex \textcolor{gray}{// mode update}
    \State Check entry flag by LiDAR scan $\mathbf{z}_k$
    \State Check exit flag by $v_{Ez,k}^{W}$ in $\hat{\mathbf{x}}_k$

    \If{$\mathrm{Flag}_{\mathrm{Exit}}$}
        \Statex \textcolor{gray}{// Exit update}
        \State Apply exit prior update via \eqref{eq:zupt}
        \State Apply re-anchoring update via \eqref{eq:reanchoring}
        \State Reset elevator-related states
    \EndIf
\EndIf

\end{algorithmic}
\end{algorithm}

\subsection{Inertial State Propagation}
\label{subsec:inertial_state_propagation}

\begin{figure}[t]
\centering
\includegraphics[width=\linewidth]{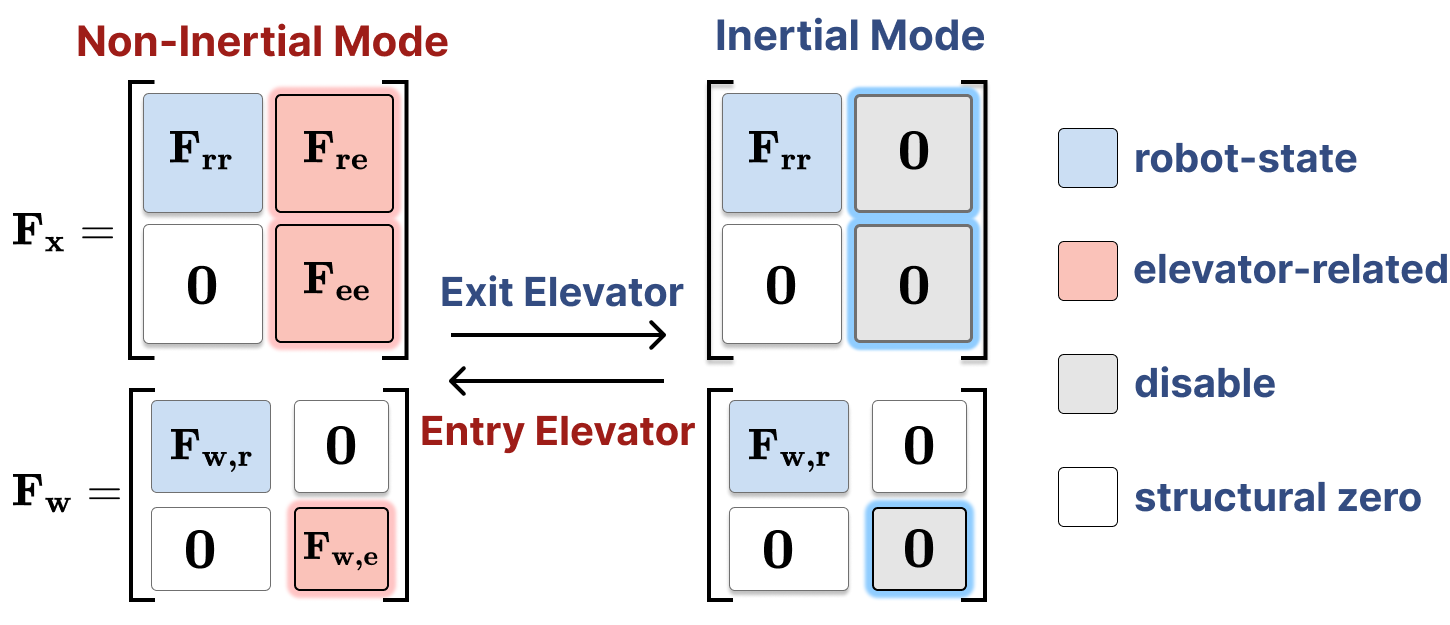}
\caption{Mode-dependent activation patterns of the error-state transition matrix $\mathbf{F}_{x,k}$ and the noise injection matrix $\mathbf{F}_{w,k}$. In non-inertial mode, the elevator-related coupling and transport-state blocks are enabled, whereas in inertial mode these elevator-related blocks are disabled. }
\label{fig:mode_dependent_propagation}
\end{figure}

When the mode manager determines that the platform is no longer inside the elevator, the estimator switches back to the standard inertial propagation model. The elevator-related blocks in the error-state transition matrix $\mathbf{F}_x$ and the noise injection matrix $\mathbf{F}_w$ are disabled, as illustrated in Fig.~\ref{fig:mode_dependent_propagation}. In particular, the elevator-to-robot cross-coupling block is set to zero, and the elevator-motion substate $[p_{Ez}^W,\, v_{Ez}^W,\, a_{Ez}^W]^T$ no longer evolves through the propagation model. The corresponding elevator-related process-noise channel is also disabled. As a result, the active part of the estimator degenerates to a standard inertial-frame LIO pipeline, while the elevator-specific states remain frozen until they are reset by the subsequent exit-handling procedure.

\subsection{State Update and Exit Constraint}
\label{subsec:state_update_and_exit_constraint}
To enable reliable operation in elevator scenarios, we introduce a tailored state-update procedure together with elevator-specific constraints. The state-estimation pipeline of Elevator-LIO is summarized in Algorithm~\ref{alg:elevator_lio}.

% \subsubsection{Outlier Rejection}

% To improve the robustness of the LiDAR update in confined elevator cabins, we reject unreliable point-to-plane constraints before residual evaluation. For each query point, a local neighborhood is first retrieved from the map, and points with insufficient support are discarded directly. A local plane is then estimated by PCA, and its physical thickness is evaluated by the root-mean-square error
% \[
% \mathrm{RMSE}=\sqrt{\lambda_{\min}/N},
% \]
% where $\lambda_{\min}$ is the smallest covariance eigenvalue and $N$ is the number of supporting points. Neighborhoods with excessively large thickness are treated as unreliable. To avoid unnecessary feature loss near useful boundaries, we further apply a one-step rescue heuristic: if the initial fit is too thick, the farthest support point is removed once and the plane is re-estimated. The recovered neighborhood is accepted only if it subsequently passes both the thickness test and a planar-structure check. Finally, the point-to-plane residual is filtered by a distance-adaptive threshold, and only the surviving points are used in the final IESKF update.

\subsubsection{IESKF Update}

For each valid point-to-plane correspondence, we construct the LiDAR residual in the elevator-local frame. Let $\mathbf{p}_j^L$ denote the $j$th valid LiDAR point in the current scan, $\mathbf{u}_j$ the normal vector of the matched local plane, and $\mathbf{q}_j$ a point on that plane. During in-elevator operation, scan-to-map registration is performed with respect to the elevator-local frame $\mathcal{F}_E$, whose geometry is attached to the cabin interior. Therefore, both $\mathbf{u}_j$ and $\mathbf{q}_j$ are expressed in $\mathcal{F}_E$.

Let
\begin{equation}
\mathbf{p}_j^I =
\mathbf{R}_L^I \mathbf{p}_j^L + \mathbf{p}_L^I
\end{equation}
be the LiDAR point transformed into the IMU frame, where $\mathbf{R}_L^I$ and $\mathbf{p}_L^I$ are the pre-calibrated LiDAR-to-IMU extrinsics. The corresponding point coordinate in the elevator-local frame is
\begin{equation}
\mathbf{p}_j^E =
\mathbf{R}_I^E \mathbf{p}_j^I + \mathbf{p}_I^E
\end{equation}
The point-to-plane residual is then defined as
\begin{equation}
\label{eq:residual}
r_j(\mathbf{x}) 
=
\mathbf{u}_j^T
\left(
\mathbf{R}_I^E
\left(
\mathbf{R}_L^I \mathbf{p}_j^L + \mathbf{p}_L^I
\right)
+
\mathbf{p}_I^E
-
\mathbf{q}_j
\right)
\end{equation}

It should be emphasized that this residual only constrains the robot state relative to the elevator-local frame. The elevator transport state does not appear in the local LiDAR residual. Instead, the global IMU position is obtained by composing the local relative position and the elevator vertical displacement:
\begin{equation}
\label{eq:global_pose_composition}
\mathbf{p}_I^W =
\mathbf{p}_I^E +
p_{Ez}^W \mathbf{e}_3
\end{equation}
Thus, during elevator traversal, the LiDAR registration remains in the local coordinate frame associated with the cabin geometry, whereas the world-frame odometry is lifted along the vertical direction by the elevator transport state.

According to the 18-dimensional error-state ordering
\begin{equation}
\delta \mathbf{x} =
\begin{bmatrix}
\delta \mathbf{p}^T,
\delta \boldsymbol{\theta}^T,
\delta \mathbf{v}^T,
\delta \mathbf{b}_a^T,
\delta \mathbf{b}_{\omega}^T,
\delta p_{Ez}^W,
\delta v_{Ez}^W,
\delta a_{Ez}^W
\end{bmatrix}^T
\end{equation}
the Jacobian of the residual with respect to the error state is
\begin{equation}
\mathbf{H}_j =
\begin{bmatrix}
\mathbf{u}_j^T &
-\mathbf{u}_j^T \mathbf{R}_I^E
\lfloor \mathbf{p}_j^I \rfloor_\wedge &
\mathbf{0}_{1\times 12}
\end{bmatrix}
\end{equation}
This zero Jacobian can be intuitively explained by the fact that, inside a fully enclosed elevator cabin, the LiDAR only observes cabin-relative geometry. Therefore, the elevator transport motion itself does not directly contribute to the LiDAR measurement residual.

This does not mean that the elevator transport states are numerically frozen during in-elevator operation. Rather, they are propagated by the IMU-driven process model, while their uncertainty gradually increases due to the lack of direct LiDAR observability. In other words, the estimator maintains a dynamically consistent prediction of the transport motion, but the corresponding covariance is expected to grow until an additional physical constraint becomes available. This is why a terminal correction at elevator exit is required.

The posterior state is estimated in an IESKF maximum-a-posteriori framework. At iteration $\kappa$, the Kalman gain and the state update are
\begin{equation}
\mathbf{K} =
\left(
\mathbf{H}^T \mathbf{R}^{-1}\mathbf{H} + \mathbf{P}^{-1}
\right)^{-1}
\mathbf{H}^T \mathbf{R}^{-1}
\end{equation}
and
\begin{equation}
\label{eq:update_x}
\hat{\mathbf{x}}^{\kappa+1} =
\hat{\mathbf{x}}^\kappa \boxplus
\left(
- \mathbf{K}\mathbf{r}(\hat{\mathbf{x}}^\kappa)
- (\mathbf{I} - \mathbf{K}\mathbf{H}) (\mathbf{J}^\kappa)^{-1}
(\hat{\mathbf{x}}^\kappa \boxminus \bar{\mathbf{x}})
\right)
\end{equation}
where $\bar{\mathbf{x}}$ and $\mathbf{P}$ are the predicted prior state and covariance, respectively, and $\mathbf{R}$ is the observation-noise covariance. The operators $\boxplus$ and $\boxminus$ denote generalized addition and subtraction on the state manifold.

Since the 18-dimensional state contains only one $\mathrm{SO}(3)$ component, namely $\mathbf{q}_I^E$, the manifold correction matrix $\mathbf{J}^\kappa$ simplifies to
\begin{equation}
\mathbf{J}^{\kappa} =
\begin{bmatrix}
\mathbf{I}_3 & \mathbf{0}_3 & \mathbf{0}_{3\times 12} \\
\mathbf{0}_3 & \mathbf{A}(\delta \boldsymbol{\theta}^\kappa)^{-T} & \mathbf{0}_{3\times 12} \\
\mathbf{0}_{12\times 3} & \mathbf{0}_{12\times 3} & \mathbf{I}_{12}
\end{bmatrix}
\end{equation}
where $\mathbf{A}(\cdot)$ is the right Jacobian of $\mathrm{SO}(3)$ and $\delta \boldsymbol{\theta}^\kappa$ is the rotational error at the current iteration. The update is repeated until convergence, yielding the posterior estimate $\hat{\mathbf{x}}_k = \hat{\mathbf{x}}^{\kappa+1}$ and covariance
\begin{equation}
\label{eq:update_P}
\hat{\mathbf{P}}_k =
(\mathbf{I} - \mathbf{K}\mathbf{H})
\mathbf{P}
(\mathbf{I} - \mathbf{K}\mathbf{H})^T
+
\mathbf{K}\mathbf{R}\mathbf{K}^T
\end{equation}
where the Joseph form is used for numerical stability.

\subsubsection{Exit Constraint and State Reset}

When the mode manager raises \texttt{Flag\_Exit}, the estimator exits the non-inertial mode and returns to the standard inertial formulation. To suppress the vertical drift accumulated during elevator traversal, we introduce an exit-stage physical-prior update together with a state re-anchoring procedure.

\paragraph{Zero velocity and zero acceleration update}

Since the elevator is physically at rest at the exit stage, the transport velocity $v_{Ez}^W$ and acceleration $a_{Ez}^W$ should approach zero. We therefore construct a high-confidence prior observation and perform one last EKF-style update. For brevity, we omit the subscript ``exit'' in this paragraph. The residual and observation matrix are
\begin{equation}
\label{eq:zupt}
\mathbf{r} =
\begin{bmatrix}
0 - v_{Ez}^W \\
0 - a_{Ez}^W
\end{bmatrix},
\quad
\mathbf{H} =
\begin{bmatrix}
\mathbf{0}_{1\times 15} & 0 & 1 & 0 \\
\mathbf{0}_{1\times 15} & 0 & 0 & 1
\end{bmatrix}
\end{equation}
The corresponding gain and correction are
\begin{equation}
\mathbf{K} =
\mathbf{P}\mathbf{H}^T
\left(
\mathbf{H}\mathbf{P}\mathbf{H}^T + \mathbf{R}
\right)^{-1},
\quad
\delta \mathbf{x} = \mathbf{K}\mathbf{r}
\end{equation}
followed by
\begin{equation}
\mathbf{x} \leftarrow \mathbf{x} \boxplus \delta \mathbf{x},
\end{equation}
\begin{equation}
\mathbf{P} \leftarrow
(\mathbf{I}-\mathbf{K}\mathbf{H})
\mathbf{P}
(\mathbf{I}-\mathbf{K}\mathbf{H})^T
+
\mathbf{K}\mathbf{R}\mathbf{K}^T
\end{equation}
In implementation, we set a small observation-noise covariance, e.g.,
$\mathbf{R}=\mathrm{diag}(10^{-5},10^{-4})$, to keep the update numerically stable while strongly suppressing residual vertical drift through the cross-covariance structure in $\mathbf{P}$.

\paragraph{Re-anchoring and transport-state absorption}

After the zero-velocity and zero-acceleration update, the elevator displacement is absorbed into the relative position state. This operation is a state re-parameterization that preserves the global position while removing the auxiliary elevator transport state. Before reset, the global IMU position is represented as
\begin{equation}
\mathbf{p}_I^W =
\mathbf{p}_I^E +
p_{Ez}^W \mathbf{e}_3
\end{equation}
To return to the standard inertial-frame LIO formulation, we re-anchor the relative position state as
\begin{equation}
\label{eq:reanchoring}
\mathbf{p}_I^E
\leftarrow
\mathbf{p}_I^E
+
p_{Ez}^W \mathbf{e}_3
\end{equation}
and then reset the elevator transport states:
\begin{equation}
p_{Ez}^W \leftarrow 0,
\qquad
v_{Ez}^W \leftarrow 0,
\qquad
a_{Ez}^W \leftarrow 0 .
\end{equation}
After this re-anchoring, the state variable $\mathbf{p}_I^E$ is again interpreted as the IMU position in the inertial world frame for subsequent standard LIO propagation. The global position is therefore kept unchanged across the reset.

\paragraph{Covariance decoupling and system reset}
\label{para:sys_reset}

After the re-anchoring step, the elevator-specific states
$[p_{Ez}^W, v_{Ez}^W, a_{Ez}^W]^T$
have completed their role and are reset to zero. To prevent the uncertainty accumulated during the non-inertial phase from contaminating subsequent conventional navigation, the corresponding cross-covariance blocks in $\mathbf{P}$ are zeroed, and the associated diagonal entries are reset to small prior values. This reset should be understood as the final step of the state re-parameterization: the elevator transport displacement has already been absorbed into the robot position state, while the auxiliary transport variables are removed before the estimator resumes standard inertial-frame operation.

\subsection{Elevator Mode Manager}
\label{subsec:elevator_mode_manager}

\begin{figure}[t]
\centering
\includegraphics[width=\linewidth]{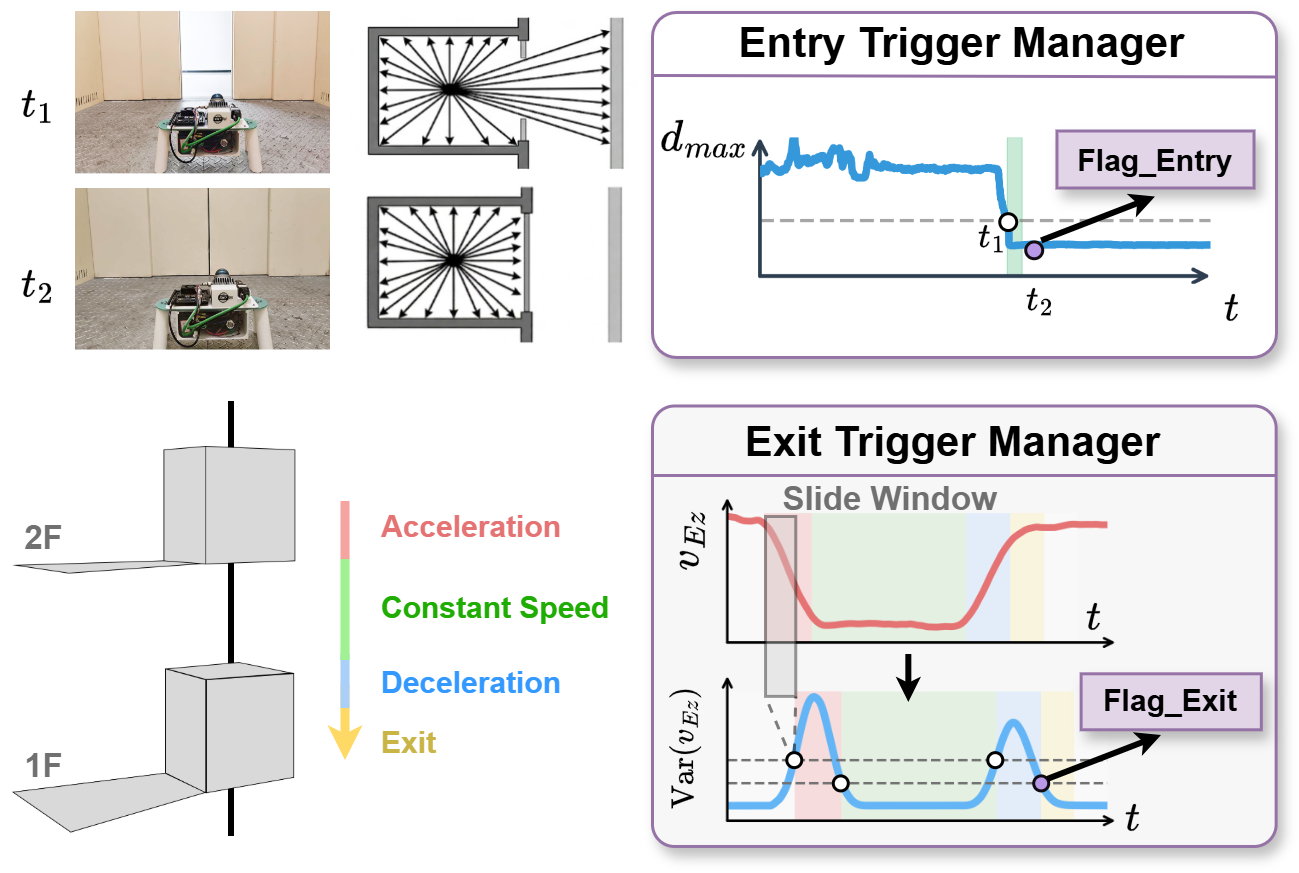}
\caption{Entry and exit trigger managers for elevator mode switching. \texttt{Flag\_Entry} is triggered by the sharp drop of the LiDAR depth metric, and \texttt{Flag\_Exit} is triggered by the variance pattern of the vertical motion during an elevator transition.}
\label{fig:ModeManager}
\end{figure}

The elevator mode manager provides the switching signals required by the core estimator to enter and leave the non-inertial mode. The implementation is intentionally modular: only the generic interfaces \texttt{Flag\_Entry} and \texttt{Flag\_Exit} are exposed to the estimator, while the underlying trigger logic can be replaced by alternative vision-based or multimodal detectors if desired. To avoid rollback while preserving the full non-inertial acceleration profile, the system should switch to the non-inertial model shortly before the elevator starts moving.

\subsubsection{Entry Trigger Based on Spatial Confinement}
An elevator has a very distinctive geometric convergence pattern. As the robot drives into the cabin, the surrounding space first becomes narrow in several directions; once the door closes, the last open direction is abruptly blocked as well. Based on this observation, we design an entry detector using the robust maximum depth of the LiDAR point cloud.

To improve robustness against noise and dynamic occlusions, we use the 94th-percentile horizontal distance of all valid LiDAR points as the reference metric, denoted by $d_{\max}$. The entry condition is defined as:
\begin{equation}
    \mathcal{F}_{\mathrm{entry}} = 
    \begin{cases} 
        1, & \text{if } d_{\max}(t) < d_{\mathrm{th}}, \ \forall t \in [t_0,\, t_0 + \Delta t_{\mathrm{door}}] \\ 
        0, & \text{otherwise} 
    \end{cases}
\label{eq:entry_condition}
\end{equation}
where $d_{\mathrm{th}}$ is the cabin-size threshold, set to 3.0 m in this paper, and $\Delta t_{door}$ is the confirmation duration, set to 2.0 s. Only when $d_{\max}$ remains below $d_{\mathrm{th}}$ throughout this interval do we declare that the elevator door has closed and emit \texttt{Flag\_Entry}, which triggers the switch to the non-inertial propagation model.

\subsubsection{Exit Trigger Based on Kinematic Variance}
Since exit detection allows a short confirmation delay after the elevator stops, we design the trigger directly from the motion estimate provided by LIO.

As shown in Fig.~\ref{fig:ModeManager}, an elevator run typically consists of acceleration, constant-speed motion, and deceleration. We use the estimated elevator vertical velocity $v_{Ez}^W$ and compute its variance within a sliding window of length $K$:
\begin{equation}
\sigma_{v_E}^2(t)
=
\frac{1}{K}
\sum_{i=0}^{K-1}
\left(
v_{Ez}^W(t-i)
-
\bar{v}_{Ez}^W(t)
\right)^2 ,
\end{equation}
where $\bar{v}_{Ez}^W(t)$ is the mean elevator vertical velocity within the current sliding window:
\begin{equation}
\bar{v}_{Ez}^W(t)
=
\frac{1}{K}
\sum_{i=0}^{K-1}
v_{Ez}^W(t-i).
\end{equation}
The variance usually exhibits two peaks during acceleration and deceleration, with a low-variance interval in between corresponding to approximately constant-speed motion. Based on this pattern, we use a finite-state machine to track the elevator cycle and declare the stop state once $\sigma_{v_E}^2(t)$ and $\left|v_{Ez}^W(t)\right|$ remain below predefined thresholds for a confirmation period after the two peaks.

% 1) \emph{Accelerating}: when the absolute z-axis velocity $|v_z|$ exceeds a motion threshold $v_{th}$ and the sliding-window variance exceeds a high threshold, $\sigma_{v_z}^2 > \sigma_{high}^2$, the system enters the accelerating state and captures the first variance peak.

% 2) \emph{Constant speed}: once the elevator reaches cruising motion, $|v_z|$ remains high but the velocity derivative approaches zero, and the variance drops below a low threshold, $\sigma_{v_z}^2 < \sigma_{low}^2$, causing a transition to the constant-speed state.

% 3) \emph{Decelerating}: when the elevator approaches the destination floor and starts braking, the velocity changes sharply again and the system detects a second variance peak, $\sigma_{v_z}^2 > \sigma_{high}^2$, thereby entering the decelerating state.

% 4) \emph{Stop check and exit confirmation}: after deceleration, the variance drops once more. To avoid false triggers caused by cabin vibration or human motion, a confirmation period $T_{confirm}$ is introduced. Only if the variance remains below $\sigma_{low}^2$ and $|v_z|$ stays near zero for longer than $T_{confirm}$ is the elevator declared completely stationary.

Once this condition is met, the finite-state machine resets and emits \texttt{Flag\_Exit}. This signal activates the exit-update module, injects the stationary prior, and returns the robot safely to standard LIO mode with a clean state.

\subsection{Static Initialization}
\label{subsec:static_initialization}

At startup, Elevator-LIO performs a short static initialization stage using the first 100 IMU measurements by default, which corresponds to about 0.5~s at 200~Hz. The average accelerometer output and gyroscope bias are computed as
\begin{equation}
\bar{\mathbf{a}}=\frac{1}{N}\sum_{k=1}^{N}\mathbf{a}_k,
\qquad
\mathbf{b}_{\omega_0}=\frac{1}{N}\sum_{k=1}^{N}\boldsymbol{\omega}_k
\end{equation}
Let $\mathcal{F}_W$ and $\mathcal{F}_I$ denote the world and IMU frames, respectively. The world z-axis is defined upward, such that $\mathbf{g}^W=[0,0,-g]^T$. The initial attitude is then obtained by aligning the normalized direction $-\bar{\mathbf{a}}/\|\bar{\mathbf{a}}\|$ with $\mathbf{g}^W/\|\mathbf{g}^W\|$. After this alignment, the remaining state variables are initialized to zero, and the gravity state is assigned $\mathbf{g}^W$. Since $\mathbf{g}^W$ is fixed after initialization and not updated during estimation, the state vector does not include a gravity state: 
\begin{equation}
\mathbf{x}_0=
\begin{bmatrix}
\mathbf{0}^T &
(\mathbf{q}_I^W)^T &
\mathbf{0}^T &
\mathbf{0}^T &
\mathbf{b}_{\omega_0}^T &
0 & 0 & 0
\end{bmatrix}^T
\end{equation}

\subsection{Geometry-Aware Adaptive Downsampling}
\label{subsec:adaptive_downsampling}

To provide stable front-end observations for the estimator, Elevator-LIO adopts an adaptive downsampling strategy instead of using a fixed-resolution voxel grid. This design is motivated by the large scene-scale variations encountered in typical deployment, where the robot repeatedly moves between spacious areas (e.g., lobbies or corridors) and highly confined elevator cabins. Under such conditions, a fixed voxel size leads to a clear resolution mismatch: it excessively removes geometric details in narrow spaces, while retaining too many redundant points in open areas.

To address this issue, we regulate the voxel size online such that the number of effective downsampled points in the current frame, denoted by $N_t$, remains close to a target value $N_{\text{target}}$. Specifically, the voxel size $v$ is updated through a lightweight proportional feedback law:
\begin{equation}
v_{t+1} = v_t \left(\frac{N_t}{N_{\text{target}}}\right)^{\frac{1}{\alpha}}
\end{equation}
where $\alpha$ is an empirical effective-dimension parameter.

This form is motivated by the non-linear relationship between point count and voxel resolution. When points are uniformly distributed in space, the number of retained points approximately scales as $N \propto v^{-3}$ for volumetric distributions and $N \propto v^{-2}$ for planar structures, while sparse LiDAR beam sampling may further reduce the effective exponent. Since real scenes are typically composed of mixed geometric structures, we introduce $\alpha$ to capture this aggregate scaling behavior in a simple yet robust manner. In our implementation, $\alpha=1.2$ is used empirically.

For robustness under abrupt viewpoint and scene changes, especially during elevator entry and exit, the updated voxel size is clamped to
\begin{equation}
v_{t+1} \in [v_{\min}, v_{\max}]
\end{equation}
where $v_{\min}=0.05$~m and $v_{\max}=0.8$~m in our implementation.

By stabilizing the effective point count across large scene-scale changes, this adaptive strategy maintains informative observations for both IESKF updates and incremental mapping, while avoiding unnecessary computation in open environments.

\subsection{Mapping}
\label{subsec:mapping}
As the downstream module of the state estimator, the mapping process registers valid points from the current LiDAR scan into the map once the IESKF update converges. To maintain a favorable balance between memory efficiency and nearest-neighbor search speed in large-scale multi-floor environments, we employ the incremental k-d tree (ikd-Tree)~\cite{cai2021ikd} as our core spatial data structure.

For each downsampled LiDAR point $\mathbf{p}_j^L$ in the current scan, it is first transformed into the IMU frame and then represented in the elevator-local frame $\mathcal{F}_E$ for local LiDAR registration:
\begin{equation}
\mathbf{p}_j^E =
\mathbf{R}_I^E \left(\mathbf{R}_L^I \mathbf{p}_j^L + \mathbf{p}_L^I\right)
+ \mathbf{p}_I^E
\end{equation}
This local representation is consistent with the point-to-plane residual used in the IESKF update and does not directly depend on the elevator transport state.

For global map insertion and odometry publication, the point is further projected into the world frame $\mathcal{F}_W$ by using the elevator's global height estimate $p_{Ez}^W$:
\begin{equation}
\mathbf{p}_j^W =
\mathbf{R}_I^E \left(\mathbf{R}_L^I \mathbf{p}_j^L + \mathbf{p}_L^I\right)
+ \mathbf{p}_I^E
+ p_{Ez}^W \mathbf{e}_3
\end{equation}
where $\mathbf{R}_L^I$ and $\mathbf{p}_L^I$ denote the calibrated LiDAR-to-IMU extrinsics, and $\mathbf{e}_3 = [0,0,1]^T$ is the unit vector mapping the scalar height to a 3-D displacement. The transformed point set $\{\mathbf{p}_j^W\}$ is then incrementally inserted into the ikd-Tree to maintain an up-to-date global map.

\section{Experiments}

In this section, Elevator-LIO is evaluated on self-collected real-world elevator datasets, public inertial benchmarks, and simulation datasets. Since the main focus of this work is LiDAR--inertial localization under elevator-induced non-inertial motion, we first validate the proposed method in real-world elevator scenarios, then assess whether it preserves competitive performance in standard inertial environments. Finally, supplementary simulation experiments are conducted to analyze its sensitivity to controlled model deviations.

\begin{figure}[t]
\centering
\includegraphics[width=\linewidth]{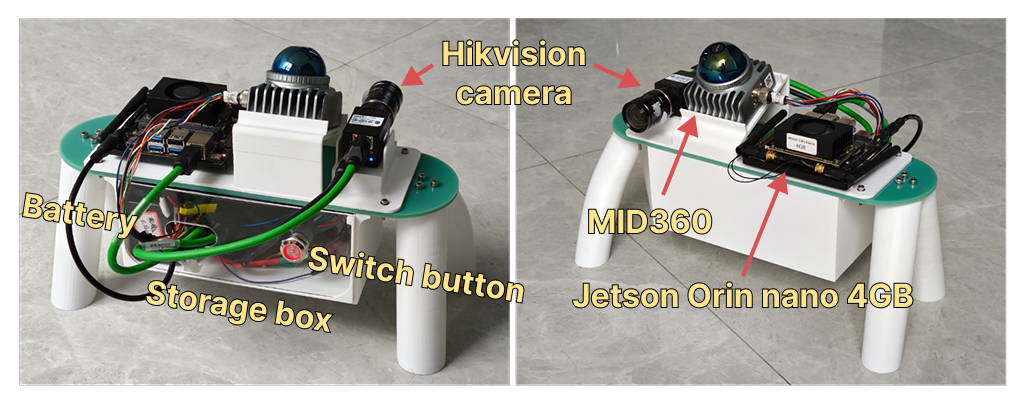}
\caption{Synchronized handheld platform used for constructing the real-world elevator dataset. The platform records LiDAR--inertial measurements and visual context for reproducible evaluation of cross-floor elevator localization.}
\label{fig:handheld}
\end{figure}

Since existing public datasets contain few dedicated sequences for cross-floor elevator operation, we construct a real-world elevator dataset covering Office, Dormitory, Campus, and Mall scenarios. As shown in Fig.~\ref{fig:handheld}, data are collected using a synchronized handheld platform equipped with a Mid360 LiDAR, a Jetson Orin Nano, and an industrial camera. The platform records LiDAR--inertial data together with synchronized visual context, enabling reproducible evaluation under elevator travel, long vertical motion, pedestrian interference, and mirror reflections. The dataset, hardware design, and processing tools will be released publicly.

The Office category includes both handheld and vehicle-mounted data acquisition, and is mainly used to evaluate robustness against additional in-cabin motion, local dynamic disturbances, and relatively complex indoor layouts. The Dormitory category typically involves larger floor spans and longer vertical round-trip motion, and is used to evaluate vertical consistency during long elevator traversals. The Campus category is characterized by long trajectories, repeated floor transitions, and large mapping ranges, making it suitable for assessing robustness and global consistency in large-scale scenes. The Mall category corresponds to open public environments with denser pedestrian traffic and higher uncertainty, and is therefore used to test adaptability in challenging real-world scenarios.

For conventional inertial environments, we use the Hilti 2022 and Hilti 2023 datasets, which cover representative indoor and outdoor scenes and are widely used for evaluating LiDAR--inertial odometry systems. In this work, they are mainly used to verify that Elevator-LIO preserves competitive localization performance in non-elevator scenarios.

As supplementary experiments, we further use simulation datasets generated by our self-developed simulator to analyze the influence of controlled model deviations, including initialization errors, cabin-attitude perturbations, and time-varying gravity-axis drift.

As baselines, we select Point-LIO, FAST-LIO2, VoxelMap, and LIO-SAM for comparison. These methods represent several typical LiDAR--inertial localization paradigms and provide a broad reference for mainstream inertial-frame LIO methods in elevator-induced non-inertial scenarios.

At the implementation level, necessary input adaptation is performed to ensure a fair comparison. Since Point-LIO and VoxelMap do not natively support the custom message format of the Mid360 sensor, we use an additional conversion node to transform the Mid360 output into standard ROS point-cloud messages. Since the original LIO-SAM does not support a 6-axis IMU or solid-state LiDAR, we employ a community-adapted version for testing. Apart from these necessary interface adaptations, all baseline methods are kept as close as possible to their original configurations and intended usage.

\definecolor{SuccColor}{HTML}{2F6F5E}   % muted teal green
\definecolor{TypeIColor}{HTML}{B87924}  % muted ochre
\definecolor{TypeIIColor}{HTML}{9C3A3A} % muted brick red
\definecolor{ZeroColor}{HTML}{8A8A8A}   % soft gray

\newcommand{\cnum}[2]{%
  \ifnum#1=0
    \textcolor{ZeroColor}{0}%
  \else
    \textcolor{#2}{#1}%
  \fi
}

\newcommand{\stat}[3]{%
  \cnum{#1}{SuccColor}/%
  \cnum{#2}{TypeIColor}/%
  \cnum{#3}{TypeIIColor}%
}

\begin{table}[t]
\centering
\caption{Failure-mode statistics in real elevator scenarios}
\label{tab:failure_summary}
\begin{tabular*}{\columnwidth}{@{\extracolsep{\fill}}lccccc}
\toprule
\textbf{Scenario} 
& \makecell{\textbf{FAST-}\\\textbf{LIO2}} 
& \makecell{\textbf{LIO-}\\\textbf{SAM}} 
& \makecell{\textbf{Point-}\\\textbf{LIO}} 
& \makecell{\textbf{Voxel-}\\\textbf{Map}} 
& \makecell{\textbf{Elevator-}\\\textbf{LIO}} \\
\midrule
Office    
& \stat{0}{9}{1}   
& \stat{0}{0}{10} 
& \stat{0}{10}{0} 
& \stat{0}{5}{5}  
& \textbf{\stat{10}{0}{0}} \\

Dormitory 
& \stat{0}{0}{4}  
& \stat{0}{0}{4}  
& \stat{0}{4}{0}  
& \stat{0}{3}{1}  
& \textbf{\stat{4}{0}{0}} \\

Campus    
& \stat{0}{2}{2}  
& \stat{0}{0}{4}  
& \stat{0}{4}{0}  
& \stat{0}{4}{0}  
& \textbf{\stat{4}{0}{0}} \\

Mall      
& \stat{0}{0}{2}  
& \stat{0}{0}{2}  
& \stat{0}{2}{0}  
& \stat{0}{2}{0}  
& \textbf{\stat{2}{0}{0}} \\

\midrule
\textbf{Overall} 
& \stat{0}{11}{9} 
& \stat{0}{0}{20} 
& \stat{0}{20}{0} 
& \stat{0}{14}{6} 
& \textbf{\stat{20}{0}{0}} \\
\bottomrule
\end{tabular*}
\par\vspace{4pt}
{\footnotesize \raggedright 
\textit{Note:} Results are reported as 
\textbf{Success} / 
\textbf{Type-I Failure} (Drift) / 
\textbf{Type-II Failure} (Divergence). \par}
\end{table}

\subsection{Experimental Setup}

Elevator-LIO is implemented in C++ and deployed on both ROS1 and ROS2. The system natively supports standard ROS point-cloud messages as well as the Livox custom message format. All real-world elevator sequences are evaluated using only a single Mid360 LiDAR, while the public benchmark datasets are tested directly using the officially released data packages. In all experiments, adaptive voxelization is enabled, with the target point number set to 20000 per second, the exponent factor set to 1.2, the minimum voxel size set to 0.05\,m, the maximum voxel size set to 0.8\,m, and the ikd-Tree voxel resolution set to 0.1\,m.

For elevator scenarios, the primary metric is the terminal vertical error, since the main concern of this work is the preservation of vertical consistency under elevator-induced non-inertial motion. Because complete trajectory-level ground truth is unavailable for the real-world sequences, we evaluate vertical consistency using same-floor revisit or stair--elevator cross-validation constraints. The terminal vertical error is defined as
\begin{equation}
e_z = z_{\mathrm{end}} - z_{\mathrm{ref}} ,
\end{equation}
where $z_{\mathrm{end}}$ is the estimated terminal height and $z_{\mathrm{ref}}$ is the reference height of the corresponding physical floor. For return-to-origin or same-floor revisit sequences, we set $z_{\mathrm{ref}}=0$, so the metric becomes $e_z=z_{\mathrm{end}}$. In several sequences, the robot reaches the same floor through both elevator and stair routes, providing cross-validation for the estimated vertical displacement.

In addition to localization accuracy, we also report the detection success rate of the elevator mode manager over elevator segments. For the very small number of sequences in which automatic state detection fails, mode switching is triggered manually to decouple estimator performance from detector failures; we analyze the causes of these detection failures later in this section. the corresponding trigger topic is also recorded in the rosbag files. Unless otherwise specified, the elevator mode manager uses the default configuration described in Section~\ref{subsec:elevator_mode_manager}, with a distance threshold of 3\,m and a percentile setting of 94%.

For standard inertial scenarios, evaluation is conducted directly against the dataset ground truth. On these sequences, Elevator-LIO is compared with several representative LIO baselines. We slightly tune the parameters of VoxelMap to improve its performance under our evaluation settings, while all other methods are run with their default settings. All experiments are conducted on a desktop computer equipped with an Intel i5-14600KF CPU and 32\,GB RAM. Real-time performance is further validated on a Jetson Orin Nano 4GB platform.

\begin{figure}[t]
\centering
\includegraphics[width=\linewidth]{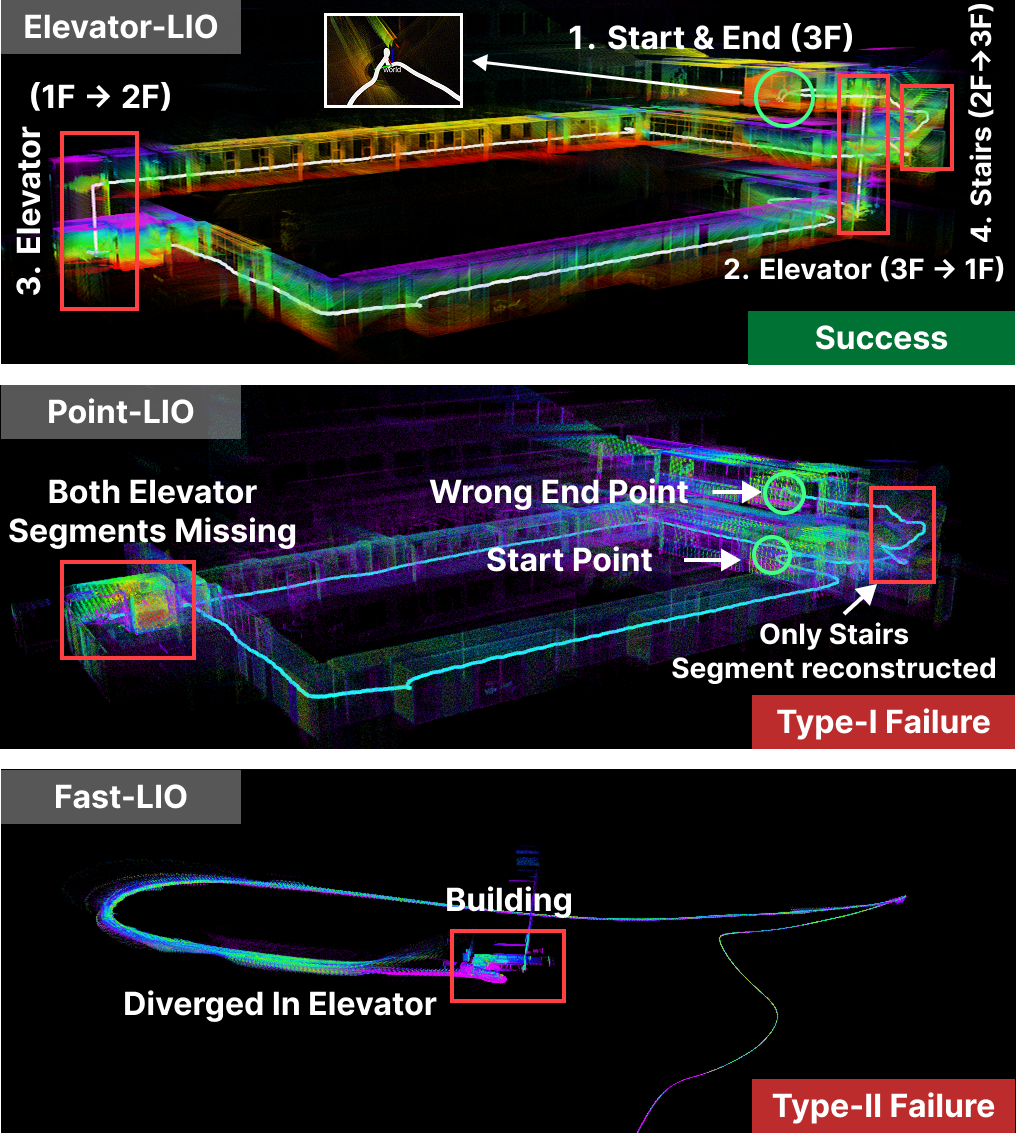}
\caption{Representative qualitative comparison in a real-world elevator scenario. The proposed Elevator-LIO achieves successful multi-floor localization. In contrast, Point-LIO shows a Type-I failure, where elevator motion is largely ignored and point clouds from different floors are incorrectly overlaid, while FAST-LIO2 shows a Type-II failure with strong pose inconsistency and eventual divergence.}
\label{fig:Failure_Modes}
\end{figure}

\newcommand{\seq}[1]{\hspace*{0.8em}#1}

\begin{table*}[t]
\centering
\caption{Terminal vertical errors of Elevator-LIO variants.}
\label{tab:ours_detailed_results}
\setlength{\tabcolsep}{3pt}
\renewcommand{\arraystretch}{1.2}
\footnotesize
\begin{tabularx}{\textwidth}{l >{\raggedright\arraybackslash}X *{7}{>{\centering\arraybackslash}X}}
\toprule
\multirow{2}{*}{\textbf{Scene}} &
\multirow{2}{*}{\textbf{Sequence}} &
\multirow{2}{*}{\makecell{\textbf{Elevator Segs.}\\\textbf{(Detected/GT)}}} &
\multirow{2}{*}{\makecell{\textbf{Elev. Ratio}\\\textbf{(\%)}}} &
\multirow{2}{*}{\makecell{\textbf{Traj. Length}\\\textbf{(m)}}} &
\multicolumn{4}{c}{\textbf{Terminal z-error (m)}} \\
\cmidrule(lr){6-9}
& & & & & \textbf{No ZUPT} & \textbf{Map Reset} & \makecell{\textbf{w/o}\\\textbf{Adapt.}} & \textbf{Full} \\
\midrule

\multirow{10}{*}{Office} 
& Office1  & 2/2 & 29.2 & 26.4  & 0.386  & 0.096  & \underline{0.010}  & \textbf{0.002} \\
& Office2  & 2/2 & 14.9 & 50.6  & 1.665  & 0.223  & \underline{-0.005} & \textbf{-0.002} \\
& Office3  & 2/2 & 3.1  & 274.5 & 0.777  & \underline{0.541}  & -6.718 & \textbf{0.001} \\
& Office4  & 2/2 & 7.3  & 109.6 & -3.074 & 0.448  & \underline{0.016}  & \textbf{-0.002} \\
& Office5  & 2/2 & 48.0 & 22.7  & -1.877 & \underline{0.218}  & 2.397  & \textbf{0.117} \\
& Office6  & 2/2 & 6.1  & 179.3 & -0.602 & \underline{-0.244} & $\times$ & \textbf{0.006} \\
& Office7  & 3/3 & 44.5 & 41.8  & -1.948 & \underline{-0.206} & -0.325 & \textbf{-0.136} \\
& Office8  & 3/3 & 97.4 & 19.1  & 0.337  & \underline{0.212}  & 1.632  & \textbf{-0.006} \\
& Office9  & 2/2 & 4.5  & 168.4 & \underline{-0.131} & 0.160 & 0.862 & \textbf{-0.002} \\
& Office10 & 3/3 & 43.5 & 42.8  & -1.877 & 0.032  & \underline{0.031}  & \textbf{-0.005} \\
\midrule

\multirow{4}{*}{Dormitory} 
& Dormitory1 & 3/3   & 83.2 & 71.9  & \underline{-0.003} & -0.112 & 0.016 & \textbf{-0.003} \\
& Dormitory2 & 8/8   & 84.4 & 92.5  & -2.310 & \underline{-0.033} & -1.857 & \textbf{-0.003} \\
& Dormitory3 & 8/8   & 85.9 & 97.5  & -0.390 & 0.017  & \underline{0.008} & \textbf{-0.004} \\
& Dormitory4 & 13/13 & 88.7 & 156.6 & -0.172 & \underline{-0.007} & $\times$ & \textbf{-0.007} \\
\midrule

\multirow{4}{*}{Campus} 
& campus1 & 2/2   & 5.7  & 243.0 & 0.739  & \underline{-0.214} & 0.501 & \textbf{-0.005} \\
& campus2 & 6/6   & 5.3  & 893.9 & -0.781 & \underline{-0.246} & $\times$ & \textbf{0.002} \\
& campus3 & 1/1   & 8.5  & 148.8 & -1.124 & \underline{-0.662} & -0.827 & \textbf{-0.499} \\
& campus4 & \textbf{1/3} & 4.0 & 315.1 & -0.989 & -0.416 & \underline{-0.002} & \textbf{0.002} \\
\midrule

\multirow{2}{*}{Mall} 
& Mall1 & \textbf{6/7} & 66.4 & 70.18 & -1.749 & -0.105 & \underline{-0.018} & \textbf{0.002} \\
& Mall2 & 5/5           & 68.9 & 67.45 & 0.738  & \underline{0.158} & 3.230 & \textbf{0.002} \\

\bottomrule
\end{tabularx}
\par\vspace{4pt}
{\footnotesize \raggedright
\textit{Note:} \textit{No ZUPT} denotes the variant without the exit-time zero-state update; \textit{Map Reset} clears and rebuilds the ikd-Tree after arriving at each new floor; \textit{w/o Adapt.} disables the adaptive front-end and uses a fixed voxel resolution of 0.2\,m; \textit{Full} denotes the complete Elevator-LIO system. Bold and underlined values indicate the best and second-best absolute terminal z-errors. The symbol $\times$ indicates that the absolute terminal z-error exceeds 10\,m.\par}
\end{table*}

\subsection{Performance in Non-Inertial Elevator Environments}

We first evaluate the applicability of mainstream LIO methods in non-inertial elevator scenarios, and the results are summarized in Table~\ref{tab:failure_summary}. Owing to the lack of dedicated modeling for non-inertial environments such as elevator traversal, all four representative baseline methods exhibit evident failures across the tested scenarios. We categorize their failure behaviors into two types: Type-I failure, corresponding to estimation drift, and Type-II failure, corresponding to system divergence. All statistics in the table are reported in the format of Success / Type-I Failure / Type-II Failure.

Type-I failure refers to an erroneously quasi-static pose estimate during elevator traversal. When the non-inertial motion is insufficient to break the LiDAR-dominant solution, conventional methods tend to prioritize geometric registration and fail to absorb the vertical motion information from the IMU. This trapping phenomenon is evident in the mild-acceleration Office sequences and is particularly common in Point-LIO, which strongly relies on LiDAR priors. Consequently, point clouds from different floors are incorrectly overlaid onto the same map without immediate divergence.

Conversely, Type-II failure is characterized by severe IMU--LiDAR observation conflicts that incorrectly drive the state update, resulting in abrupt pose jumps and eventual divergence. This typically occurs under stronger non-inertial excitation, such as the higher acceleration in the Dormitory sequences that causes FAST-LIO2 to fail. Fundamentally, this reflects the inability of conventional LIO frameworks to simultaneously satisfy both inertial and geometric constraints under elevator-induced non-inertial motion. Fig.~\ref{fig:Failure_Modes} presents representative qualitative comparisons of these failure modes.

\begin{table*}[t]
\caption{Absolute Translational Errors (RMSE, meters) on Hilti Dataset}
\label{tab:ape_comparison}
\centering
\resizebox{\textwidth}{!}{
\begin{tabular}{llcccccc}
\toprule
Dataset & Sequence & Faster-
LIO\cite{fasterlio} & LIO-SAM\cite{lio_sam} & Point-LIO\cite{pointlio} & VoxelMap\cite{Voxelmap} & FAST-LIO2\cite{fast_lio2} & Elevator-LIO (Ours) \\
\midrule
\multirow{8}{*}{Hilti2023}
& Floor 0 & \underline{0.016} & 0.046 & \textbf{0.011} & 0.017 & 0.023 & 0.024 \\
& Floor 1 & 0.028 & $\times$ & 0.025 & 0.058 & \textbf{0.008} & \underline{0.017} \\
& Floor 2 & \textbf{0.019} & 0.265 & 0.087 & 0.077 & \underline{0.046} & 0.107 \\
& Stair & 0.864 & $\times$ & \underline{0.173} & $\times$ & \textbf{0.124} & 1.262 \\
& Underground 1& \underline{0.019} & 0.080 & 0.041 & 0.068 & 0.031 & \textbf{0.013} \\
& Underground 2 & 0.167 & $\times$ & 0.106 & \textbf{0.049} & 0.125 & \underline{0.084} \\
& Underground 3 & 0.117 & $\times$ & 0.108 & \underline{0.075} & \textbf{0.038} &  0.106 \\
& Underground 4 & \underline{0.022} & 0.161 & 0.060 & 0.036 & 0.034 & \textbf{0.019} \\
\midrule
\multirow{9}{*}{Hilti2022}
& 01 Construction Ground level & \textbf{0.012} & 0.078 & 0.019 & 1.827 & \underline{0.015} & 0.016 \\
& 02 Construction Multilevel  & 0.037 & 1.889 & 0.059 & $\times$ & \textbf{0.026} & \underline{0.033} \\
& 03 Construction Stairs & $\times$ & $\times$ & 3.070 & $\times$ &  0.446 & \textbf{0.331} \\
& 07 Long Corridor & \underline{0.053} & $\times$ & \textbf{0.043} & 0.073 & 0.071 &  1.019 \\
& 11 Lower Gallery & \underline{0.810} & 0.834 & 1.504 & 10.337 & 1.207 & \textbf{0.744} \\
& 15 Attic to Upper Gallery & $\times$ & $\times$ & \textbf{1.526} & $\times$ & $\times$ & \underline{2.873} \\
& 21 Outside Building & \underline{0.051} & 3.061 & 0.134 & 23.300 & 0.248 & \textbf{0.029} \\
& 06  Constr. Upper Level 2 & \underline{0.034} & 0.085 & \textbf{0.030} & 1.288 &  0.041 &  0.047 \\
& 14 Basement 2 & 0.166 & 33.723 & 0.107 & 0.622 & \underline{0.064} & \textbf{0.063} \\
\bottomrule
\end{tabular}
}
\end{table*}

We then conduct a quantitative evaluation of Elevator-LIO and several ablated variants in elevator scenarios. Since this work primarily focuses on the state-estimation capability of the system under elevator-induced non-inertial motion, the terminal vertical error is adopted as the main evaluation metric, and all errors are reported in meters. In addition, we report the detection performance of the elevator mode manager by counting the number of automatically detected elevator segments over the annotated ground-truth elevator segments. The corresponding results are summarized in Table~\ref{tab:ours_detailed_results}, where the definitions of the ablated variants and table entries are provided in the table note.

\begin{figure*}
\centering
\includegraphics[width=0.95\linewidth]{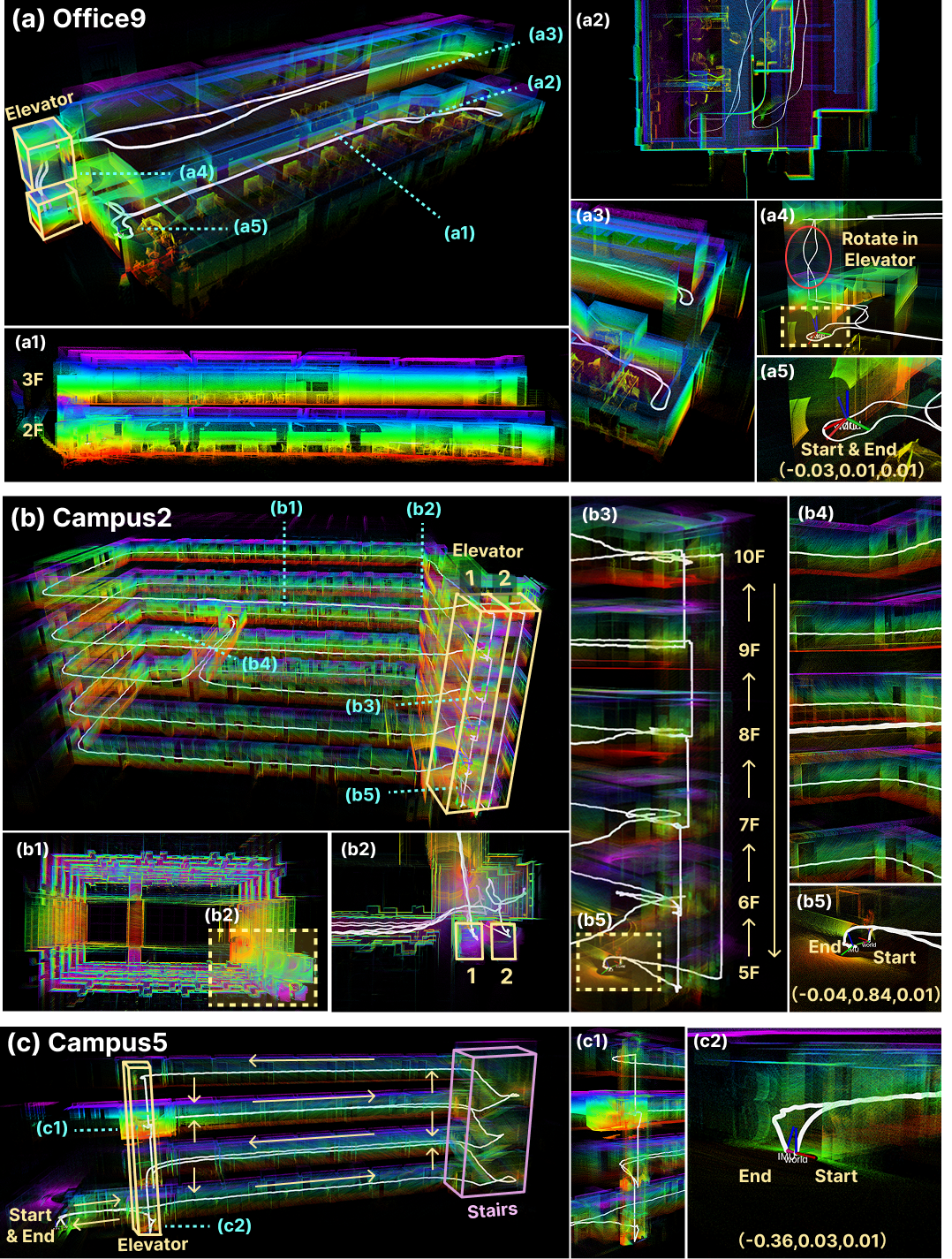}
\caption{Representative large-scale real-world elevator sequences in the experimental dataset. This category emphasizes cross-floor mapping consistency over long trajectories in large multi-floor environments. Insets show representative elevator regions and local trajectory details. The numbers in the lower-right corners indicate the final world-frame positions of the corresponding trajectories.}
\label{fig:Scenarios}
\end{figure*}

The ablation results show that the exit-time \textit{zero-state update} is critical for suppressing vertical drift. Without this correction, accumulated height errors can reach meter scale in several sequences, whereas the full system keeps the terminal height error below 1\,cm in most cases. 

The \textit{Map Reset} variant removes the correction effect from historical-map revisits and therefore provides a more practical evaluation of the combined effect of forward non-inertial prediction and exit-time zero-state correction after each elevator transition. The millimeter-level terminal errors achieved by the full system should be interpreted as the combined outcome of local non-inertial estimation and historical-map alignment, rather than as evidence that sub-centimeter absolute height accuracy is maintained throughout the entire elevator traversal.

The \textit{w/o Adapt.} variant further demonstrates the importance of the adaptive front-end for system stability. When the adaptive front-end is disabled and a fixed voxel resolution is used, the estimator becomes more sensitive to abrupt scene-scale changes between open areas and confined elevator cabins. As a result, the terminal vertical error increases significantly in several sequences, and some cases even exceed the valid error range. This indicates that maintaining an appropriate point-cloud resolution is critical for preserving effective geometric constraints during elevator entry, in-cabin operation, and cross-floor transitions.

\begin{figure}[t]
\centering
\includegraphics[width=0.95\linewidth]{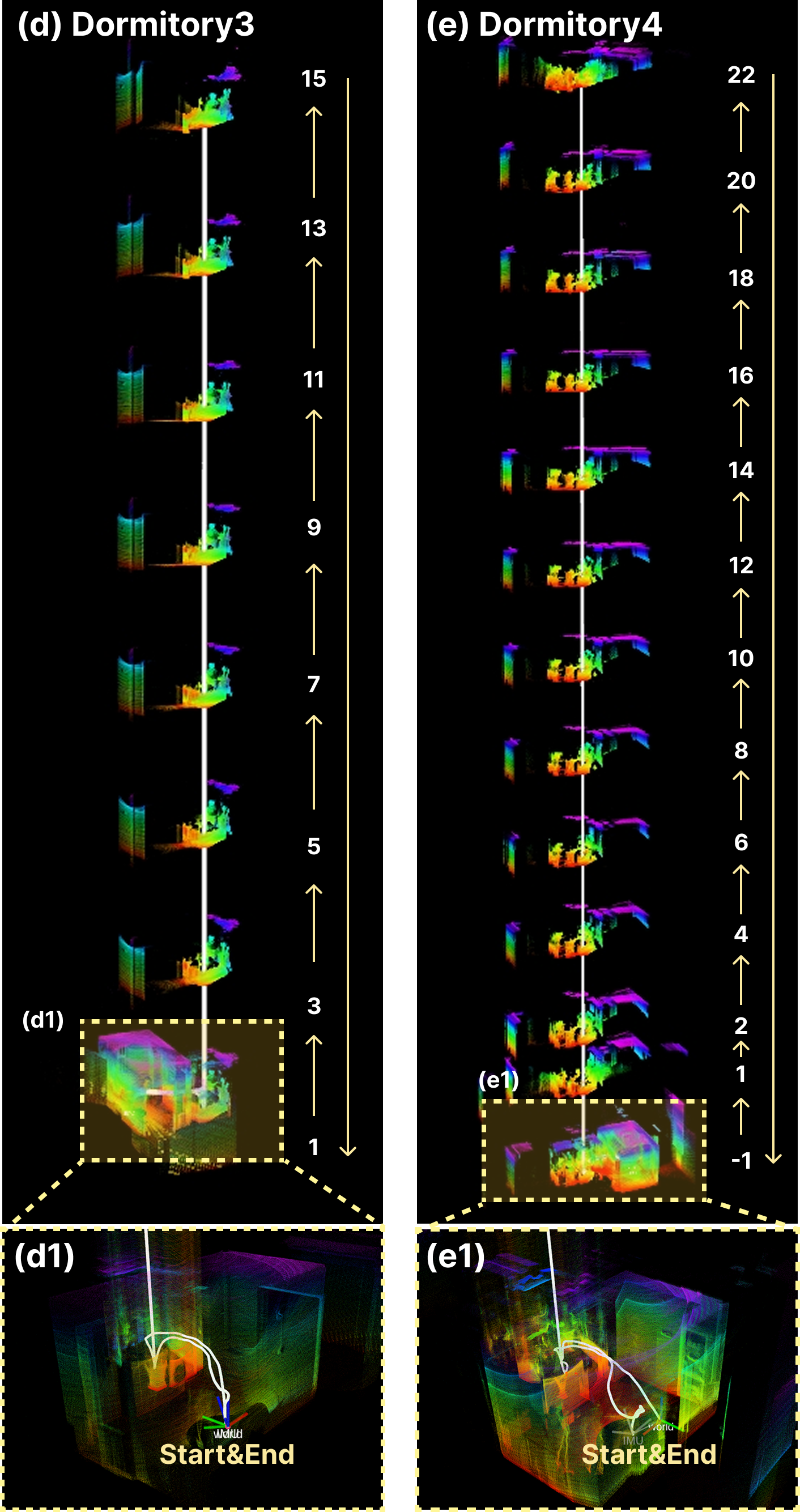}
\caption{Representative long-range vertical traversal in the Dormitory sequences. These sequences evaluate vertical consistency during repeated floor-by-floor elevator travel and direct return to the starting floor.}
\label{fig:Long_Scenarios}
\end{figure}

The tested sequences cover two dominant difficulties, as illustrated in Fig.~\ref{fig:Scenarios}. Dormitory and Mall sequences have high elevator ratios and mainly stress long-range vertical consistency under sustained elevator motion. In particular, the Dormitory sequences evaluate repeated floor-by-floor traversal and direct return to the starting floor, as shown in Fig.~\ref{fig:Long_Scenarios}. For example, Dormitory4 accumulates 138.9\,m of vertical travel and still returns close to the initial height. In contrast, Office and Campus sequences contain longer horizontal trajectories, repeated floor transitions, and larger mapping ranges, stressing cross-floor map consistency and robustness to additional in-cabin motion. Elevator-LIO remains stable in both categories, including handheld sequences with strong local motion disturbances and public environments with pedestrian interference.

\begin{figure}[t]
\centering
\includegraphics[width=\linewidth]{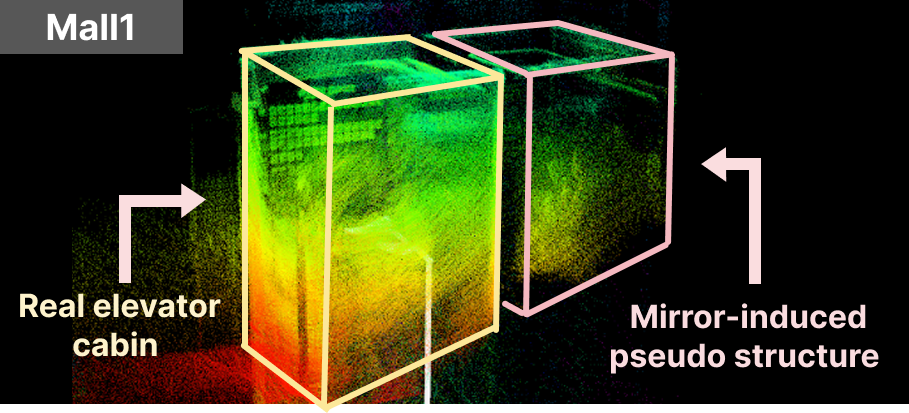}
\caption{A representative failure case of elevator-entry detection in a mirrored cabin. Reflection and multipath effects generate pseudo structures in the point cloud, making a single elevator appear as multiple adjacent cabins, which may cause the distance-threshold-based detector to fail in rare cases. Second-order and even higher-order reflected structures can also be observed in the background.}
\label{fig:reflect}
\end{figure}

Beyond localization accuracy, we also evaluate the performance of the elevator mode detector. Among the 79 annotated elevator segments, the detector successfully identifies 76 segments. The occasional detection failures mainly occur in Campus4 and Mall1, where mirror reflections and multipath effects create pseudo-structures in the point cloud, causing the distance-threshold-based detector to miss some entry events. This behavior is expected for a lightweight heuristic detector, which cannot guarantee a 100\% recognition rate in all extreme environments. More advanced vision-based, semantic, or learning-based detectors could further improve the robustness of elevator-mode triggering in such cases. For the sequences with missed detections, we manually publish the \texttt{Flag\_Entry} signal to isolate the evaluation of the core estimator from the detector failure. Fig.~\ref{fig:reflect} shows a representative mirrored-cabin example.

% All error values are reported in meters and correspond to the terminal vertical error at the nd of each sequence. 
% E-LIO (w/o ZUPT) denotes the proposed framework without the exit-time ZUPT correction. 
% E-LIO (Map Reset) denotes the variant that resets the local map after each floor arrival. 
% E-LIO denotes the full proposed system. 
% Vertical Travel denotes the total vertical travel distance of the sequence.

% This subsection evaluates the core performance of Elevator-LIO in real-world non-inertial elevator transitions. We report end-to-end trajectory accuracy, floor-level height consistency, and robustness under different cabin sizes and motion patterns. To keep the comparison physically meaningful, all ablation variants explicitly model one coupled acceleration component during elevator motion.

\subsection{Performance in Standard Inertial Environments}

We further evaluate Elevator-LIO on the Hilti 2022 and Hilti 2023 datasets. When the platform is outside an elevator, the elevator-related state blocks and process-noise channels are disabled, and the estimator follows the standard inertial propagation and LiDAR update pipeline. Therefore, this experiment examines whether the reformulated LIO subsystem retains general applicability in conventional indoor and outdoor scenarios.

For this evaluation, we report the absolute translational error RMSE in meters, as summarized in Table~\ref{tab:ape_comparison}. The same public dataset sequences and evaluation protocol are used for all compared methods. The symbol $\times$ indicates that the corresponding method fails to complete the sequence or produces an invalid trajectory for evaluation.

As shown in Table~\ref{tab:ape_comparison}, Elevator-LIO remains competitive on standard inertial benchmarks. It achieves the best or second-best performance on several sequences, including \emph{Underground 1}, \emph{Underground 4}, \emph{11 Lower Gallery}, \emph{21 Outside Building}, and \emph{14 Basement 2}, indicating that the proposed mode-dependent design preserves normal LIO behavior in ordinary inertial environments.

We also inspect the failure cases of the compared methods and find that narrow staircases and cross-floor transition areas are particularly challenging. For example, in 2022\_exp02, VoxelMap fails at around 410\,s after the platform moves from a spacious area into a narrow staircase. In the corresponding Hilti 2023 test sequence, it also fails in a narrow staircase region at around 71\,s. In exp15, multiple methods fail to pass through the transition stably. These failures are mainly caused by the abrupt shrinkage of the point-cloud distribution in confined spaces, where effective geometric constraints become weaker and repetitive structures such as stair steps, handrails, and walls can introduce scan-matching ambiguities.

In contrast, Elevator-LIO maintains a higher local resolution in confined environments through its adaptive front-end, preserving more effective geometric constraints and improving robustness in these challenging transitions. Although Elevator-LIO is not uniformly optimal on all conventional sequences, such as \emph{Stair} and \emph{07 Long Corridor}, the Hilti results suggest that the underlying LIO subsystem still maintains competitive localization accuracy in conventional scenarios, while the adaptive front-end improves robustness in confined transition areas.

\subsection{Simulation-Based Robustness Analysis}

In real-world environments, the motion state of an elevator itself is difficult to control, making it hard to systematically evaluate different types of model mismatch. We therefore use a customized elevator simulator to evaluate robustness under controlled violations of the modeling assumptions, including time-varying gravity drift, cabin-axis inclination, initial gravity misalignment, and their combined effect. The small, medium, and large perturbation levels correspond to increasing mismatch magnitudes: the gravity-drift step is set to $0.10^\circ$, $0.30^\circ$, and $0.80^\circ$; the cabin-inclination pitch is set to $0.30^\circ$, $1.00^\circ$, and $3.00^\circ$; and the initial gravity-misalignment angle is set to $0.50^\circ$, $1.50^\circ$, and $4.00^\circ$, respectively. The combined setting applies all three perturbations simultaneously at the corresponding level.

As shown in Table~\ref{tab:model_sensitivity}, larger perturbations lead to higher errors, but the error remains bounded under all tested settings. This behavior demonstrates the robustness of the proposed system: even when the real elevator motion slightly deviates from the basic modeling assumptions, the system can still operate stably without divergence.

\begin{table}[t]
\centering
\caption{Return-to-origin height error $e_{\mathrm{ret}}=|\Delta z_{\mathrm{est}}-\Delta z_{\mathrm{gt}}|$ under simulated perturbations. All values are in meters.}
\label{tab:model_sensitivity}
\begin{tabular}{lccc}
\toprule
\textbf{Perturbation Type} & \textbf{Small} & \textbf{Medium} & \textbf{Large} \\
\midrule
Time-Varying Gravity Drift   & 0.012 & 0.036 & 0.118 \\
Cabin Inclination Error      & 0.009 & 0.028 & 0.094 \\
Initial Gravity Misalignment & 0.015 & 0.051 & 0.167 \\
Combined Perturbation        & 0.024 & 0.083 & 0.312 \\
\bottomrule
\end{tabular}
\end{table}

\section{Conclusion}

In this paper, we proposed Elevator-LIO, a robust LiDAR-inertial odometry framework specifically designed to tackle the localization challenges in the typical non-inertial environment of an operating elevator. By explicitly decoupling the absolute transport motion of the elevator from the relative motion of the robot within the state estimation, and incorporating a zero-state update (including zero-velocity and zero-acceleration constraints) mechanism tailored to the "stop-and-go" kinematic characteristics of elevators, the proposed system effectively overcomes the drift and divergence issues caused by dynamic model mismatch in conventional LIO under non-inertial frames. Furthermore, a feedback-driven adaptive downsampling strategy ensures the stability of front-end geometric constraints during the drastic spatial scale transitions between open floors and confined elevator cabins. Extensive experiments on 20 real-world sequences covering various challenging conditions, alongside comprehensive simulation evaluations, demonstrate that Elevator-LIO maintains continuous and highly consistent multi-floor localization in scenarios where mainstream baseline methods fail.

Despite these promising results, several limitations remain. The current elevator state detector mainly relies on LiDAR-based geometric confinement cues, and may therefore miss detections in non-standard elevators, such as fully mirrored, transparent, or large freight cabins. Future research could integrate visual semantics, building-side IoT signals, or floor-plan priors to build a more generalizable multimodal state-management module.

In addition, the current formulation mainly models elevator-induced vertical translation, while rotational and lateral disturbances are treated as negligible. Extending this framework to more general moving non-inertial carriers, such as subways, buses, ships, and aircraft, is a promising direction for robust localization in real-world moving-base environments. As Elevator-LIO provides a reliable odometry module for cross-floor elevator traversal, integrating it with place recognition and factor-graph optimization back-ends could further enable practical long-term multi-floor navigation.

Code, datasets, and the simulator will be released to the community to facilitate further research in non-inertial robotics localization.

\bibliographystyle{IEEEtran}
\bibliography{references}

\end{document}